\documentclass[11pt]{article}

\usepackage[final]{acl}

\usepackage{xeCJK}
\usepackage{times}
\usepackage{latexsym}

\usepackage{xcolor}
\usepackage{amsmath}
\usepackage{amssymb}
\usepackage[most]{tcolorbox}
\usepackage{listings}

  \usepackage{booktabs}
\usepackage{multirow}

  \newtcblisting{promptbox}[2][]{
    enhanced,
    breakable,
    colback=gray!6,
    colframe=gray!45,
    coltitle=black,
    fonttitle=\bfseries,
    title={#2},
    boxrule=0.4pt,
    arc=1mm,
    left=1.5mm,
    right=1.5mm,
    top=1mm,
    bottom=1mm,
    listing only,
    listing options={
      basicstyle=\ttfamily\footnotesize,
      breaklines=true,
      columns=fullflexible,
      keepspaces=true
    },
    #1
  }

\usepackage[T1]{fontenc}

\usepackage[utf8]{inputenc}

\ifxetex\else
  \usepackage{microtype}
\fi

\usepackage{inconsolata}

\usepackage{graphicx}

\title{    Rubric-as-Experts: Case-Specific MQM Rubrics for Translation Error Span Detection}

\author{
  \textbf{Weilu Xu},
  \textbf{Yunzhi Shen},
  \textbf{Xinye Wang},
  \textbf{Renfei Dang},
  \textbf{Shujian Huang}\thanks{Corresponding author.}
  \\
  National Key Laboratory for Novel Software Technology, Nanjing University
  \\
  \texttt{\{weilu.xu, shenyunzhi, xinyewang, dangrf\}@smail.nju.edu.cn}
  \\
  \texttt{huangsj@nju.edu.cn}
}

\begin{document}
\maketitle
\begin{abstract}
Large language models (LLMs) have shown potential for reference-free span-level translation quality estimation (QE), yet existing approaches based on Multidimensional Quality Metrics (MQM) typically rely on fixed rubric configurations shared across translation instances. However, translation instances often differ substantially in error complexity, ambiguity, and required evaluation granularity, making static rubric allocation suboptimal for span-level error detection. We find that larger MQM subtype spaces improve error coverage but also introduce more false positives, while different translation instances prefer different rubric granularities, suggesting that evaluation spaces should be allocated dynamically for each case. Therefore, we propose a case-specific dynamic rubric framework that adaptively constructs MQM evaluation spaces for individual translation instances. Unlike methods that generate fully free-form rubrics, our framework remains grounded in the predefined MQM taxonomy while dynamically selecting suitable subtype spaces and evaluation granularity for different cases. Experiments on span-level QE benchmarks from the Conference on Machine Translation (WMT) across multiple model scales demonstrate that the proposed framework consistently improves F1 and Matthews Correlation Coefficient (MCC).
\end{abstract}
\section{Introduction}

Large language models (LLMs) have recently demonstrated strong performance in machine translation generation \cite{brown2020language, achiam2023gpt, yang2025qwen3}.
However, their gains in fine-grained translation quality evaluation (QE), particularly span-level error localization, have been less pronounced, while specialized QE systems such as COMETKiwi \cite{rei2022cometkiwi} remain highly competitive.
Given a source sentence and its machine translation, sentence-level QE assigns an overall quality score, whereas span-level QE must identify the exact erroneous spans in the translation. This finer-grained task is commonly formulated under structured criteria such as Multidimensional Quality Metrics (MQM), which organizes translation errors hierarchically into major categories and fine-grained subtypes \cite{lommel2014multidimensional, freitag2021experts}.

\begin{figure*}[t]
    \centering
    \includegraphics[width=\textwidth]{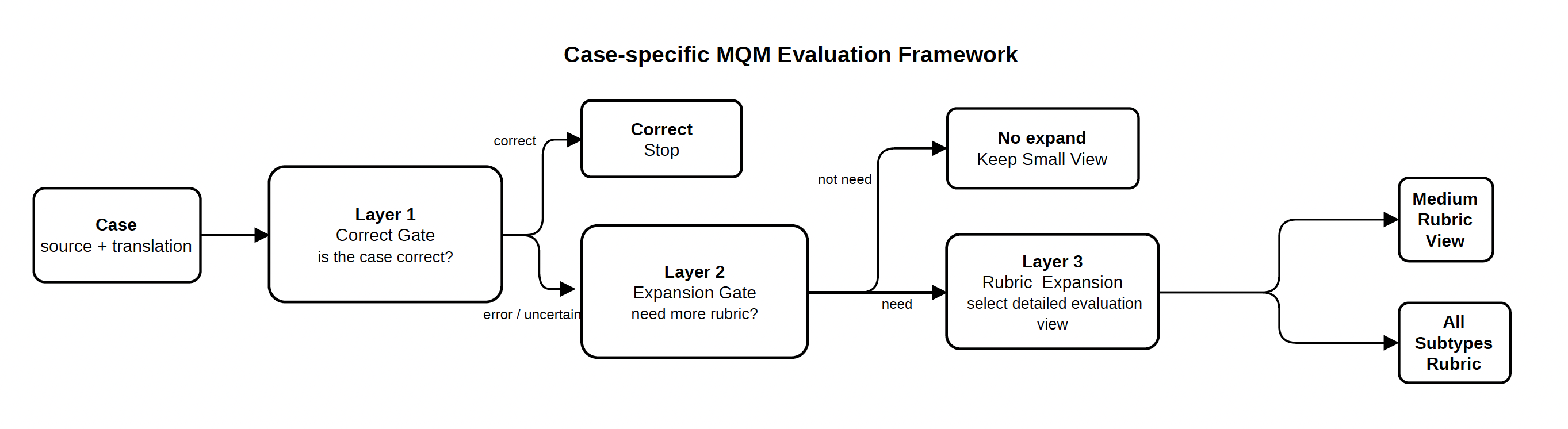}
    \caption{
    Illustration of static versus case-aware rubric exposure for span-level QE. Static rubric settings expose different translation cases to the same predefined set of MQM criteria, which may lead to insufficient error coverage or unnecessary exploration. 
    In contrast, the proposed framework selects relevant MQM categories and adjusts the number of exposed subtypes for each case, enabling a more focused evaluation search space for span-level error localization.
    }
    \label{fig:intuition}
\end{figure*}

Prior work has explored automatically generating a specific rubric for each evaluation case.
For example, \citet{fan2024sedareval} automatically generate and validate a self-adaptive rubric for each question.
MQM-based translation evaluation, however, relies on a predefined error taxonomy originally developed for systematic human annotation.
Existing span-level QE approaches use MQM as a consistent and interpretable error taxonomy, but typically expose the same predefined criteria to the evaluator for every translation instance \citep{kocmi2023gemba,kim2025rubric}.
Together, these two lines of work motivate retaining the structured MQM taxonomy while adapting the exposed evaluation criteria to individual translation instances.
This leaves open how such instance-specific adaptation should be carried out within the MQM taxonomy.

In practice, a source sentence may have multiple acceptable translations, making the distinction between legitimate variation and genuine translation error inherently ambiguous.
Within the fixed MQM taxonomy, we view the exposed criteria and their granularity as defining an evaluation search space for error identification.
Importantly, the source--translation pair itself provides observable cues for adapting this search space: lexical, numerical, terminological, or structural characteristics may indicate which MQM error categories are more relevant, while the complexity and diversity of potential error risks may indicate whether a compact or broader set of fine-grained subtypes is required.
Thus, category selection determines which error dimensions are emphasized, whereas granularity controls the breadth of fine-grained error types considered within those dimensions.
Exposing the same set of criteria at the same granularity to every instance may therefore lead to either insufficient coverage or unnecessary exploration.

Based on these observations, we propose a case-specific dynamic rubric framework for span-level QE.
For each translation instance, the framework selects relevant MQM major categories and assigns a subtype budget, which specifies how many fine-grained subtype descriptions are exposed within the selected categories and thereby controls the granularity of the rubric.
Experiments on span-level QE benchmarks show that dynamic rubric allocation consistently improves F1 and MCC over static evaluation settings.

The contributions of this work are summarized as follows:

\begin{itemize}

    \item We present a systematic analysis of how the MQM evaluation search space influences LLM-based span-level QE, showing that subtype exposure affects model exploration behavior and that the post-hoc oracle subtype budget varies substantially across instances.

    \item We propose a case-specific dynamic rubric framework that constructs an instance-specific MQM evaluation search space by selecting relevant major categories and assigning a subtype budget.

    \item Experiments on span-level QE benchmarks show that dynamic rubric allocation improves F1 and MCC.

\end{itemize}

\section{Related Work}

\paragraph{LLM-based Translation Quality Evaluation}
Translation quality evaluation (QE) has gradually evolved from sentence-level quality scoring toward fine-grained token- and span-level error detection. Instead of only predicting an overall quality score, recent QE systems aim to identify concrete error spans. MQM provides a structured taxonomy for fine-grained translation evaluation, defining hierarchical error categories such as accuracy, fluency, terminology, and style \citep{lommel2014multidimensional}. Building on this framework, recent LLM-based QE methods show that large language models can perform MQM-style analytic evaluation through prompting \citep{kocmi2023gemba}. These studies suggest that LLMs already possess relatively strong fine-grained error perception ability. However, existing methods typically rely on fixed prompts or static rubric configurations shared across all samples. As a result, the model's error exploration behavior is strongly constrained by a predefined evaluation space, even though different translation cases may require substantially different levels of error exploration and rubric granularity.

\paragraph{Rubric-conditioned Evaluation}
Rubric-based and criterion-conditioned evaluation have become increasingly important in LLM-as-a-judge research \citep{liu2023geval,anghel2025rubric}. By explicitly providing evaluation criteria in natural language, these methods guide the model's reasoning focus and evaluation behavior, often improving interpretability and controllability. Recent studies further explore automatically generated task- or instance-specific rubrics, where evaluation criteria are constructed according to the current input \citep{fan2024sedareval,siro2026learning}. Although these methods demonstrate that evaluation criteria need not remain fully static, their generated criteria are not constrained to a predefined domain taxonomy and may vary in coverage and granularity across cases. In translation quality evaluation, different cases often exhibit substantial variation in ambiguity, error density, and required exploration granularity. Free-form criteria generation may therefore lead to unstable coverage, inconsistent granularity, and uncontrolled evaluation behavior. Meanwhile, MQM-based QE methods still mainly rely on static rubric configurations, where all samples share the same subtype exploration space. In contrast, our work combines MQM’s structured subtype hierarchy with case-specific dynamic rubric allocation. Rather than generating free-form criteria, our routing framework selects MQM major categories and allocates a subtype budget for each translation instance.

\paragraph{Adaptive Routing and Dynamic Search Space}
Increasing the evaluation search space generally improves recall because the model can explore a broader range of potential error dimensions. However, exhaustive subtype exploration also tends to introduce large numbers of false positives and substantially higher inference cost. This suggests that different translation samples may require different exploration granularity: simpler cases may only need a compact subtype space, while more ambiguous or error-dense cases may benefit from broader exploration. Similar ideas have been widely studied in adaptive computation, mixture-of-experts routing, and test-time scaling, where models dynamically allocate computation budgets, reasoning paths, or expert activations according to input complexity \citep{graves2016act,shazeer2017outrageously,fedus2022switch,zhang2025survey}. In mixture-of-experts systems, a router activates a subset of parameterized experts and thereby determines which specialized computation is available for each input. We adopt an analogous view at the evaluation level: each MQM category or subtype rubric provides an expert-like evaluation perspective, and the activated rubric set operationally bounds the error dimensions that the evaluator is explicitly guided to explore. Unlike conventional parameterized experts, these rubric experts do not modify the evaluator's weights; they constrain its instance-specific evaluation search space through natural-language criteria. Related test-time learning studies further show that models can adapt at inference time to the characteristics of test data, for example by prioritizing high-perplexity samples for lightweight test-time updates \citep{hu2025test}. However, existing routing methods mainly focus on token computation, expert activation, or reasoning-budget allocation, while relatively little attention has been paid to dynamically controlling the evaluation search space itself. Our work introduces routing into MQM-based QE to dynamically allocate rubric granularity and subtype exploration space for different translation cases, enabling more adaptive and case-specific error exploration.

\section{Method}
\label{sec:method}

\subsection{Task Formulation}
\label{sec:method-task}

Given a source sentence $x$ and its machine translation $y$, the task is to
detect erroneous spans in $y$. The output is a set of text spans:
\begin{equation}
    \hat{\mathcal{S}} = \{ \hat{s}_1, \hat{s}_2, \ldots, \hat{s}_n \},
\end{equation}
where each $\hat{s}_i$ is a contiguous span copied from the translation.

Unlike sentence-level QE, span-level evaluation requires the model to localize
concrete textual evidence. The exposed evaluation criteria therefore strongly
affect the model's search behavior. Narrow rubrics may miss valid errors,
whereas overly broad rubrics often introduce noisy detections.

\begin{figure*}[t]
    \centering
    \includegraphics[width=0.95\textwidth]{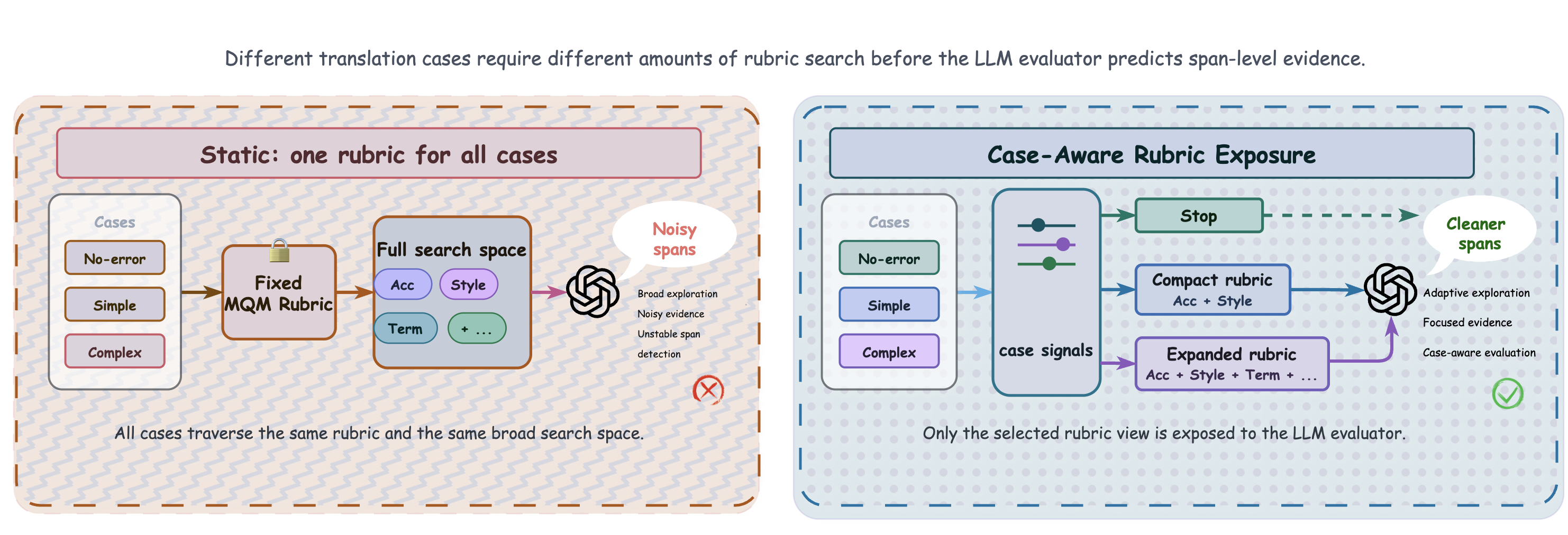}
    \caption{
    Overview of the case-specific MQM evaluation framework. The input case
    consists of a source sentence and its machine translation. The framework
    first uses a correctness gate to stop evaluation for cases predicted as
    correct. For error-prone or uncertain cases, an expansion gate determines
    whether the compact rubric view is sufficient. If further exploration is
    needed, the rubric expansion module selects a more detailed evaluation
    view, such as a medium-granularity rubric view or the full MQM subtype
    rubric view.
    }
    \label{fig:framework}
\end{figure*}

We formulate span-level QE with instance-specific construction of the evaluation search space. Instead of exposing the same MQM criteria to every translation, the routing framework first performs major-category selection and then allocates a subtype budget for each source--translation pair. Figure~\ref{fig:framework} illustrates the
overall cascaded evaluation framework. Given a source--translation pair, the
framework first applies a correctness gate to filter cases that do not require
further evaluation. For the remaining error-prone or uncertain cases, it then
decides whether compact rubric evaluation is sufficient or whether broader MQM
subtype exploration is needed. The resulting search space uses one of three subtype budgets: Compact, Expanded, or Comprehensive.

\subsection{Structured MQM Rubric Space}
\label{sec:method-rubric-space}

We construct a structured rubric space from the MQM taxonomy. Each MQM subtype
is paired with a natural-language description and treated as an atomic rubric
unit. A case-specific rubric is formed by selecting a subset of these units:
\begin{equation}
    \mathcal{R}(x,y)
    \subseteq
    \mathcal{R}_{\mathrm{MQM}},
\end{equation}
where $\mathcal{R}_{\mathrm{MQM}}$ denotes the complete MQM-derived rubric
pool.

Unlike the free-form generation of rubrics, the method does not invent new evaluation
criteria. Instead, it dynamically selects from a predefined MQM-based rubric
space, preserving MQM interpretability while adapting the exposed criteria to
different translation cases.

\subsection{Two-stage MQM Routing}
\label{sec:method-router}

Exposing every MQM subtype description to every translation is costly and can increase false positives. We therefore introduce a lightweight routing framework that first performs MQM major-category selection and then selects a subtype budget.

\paragraph{Major-category router.}
Given $(x,y)$, we encode the source--translation pair using a Qwen3 causal
language model and use the final hidden state as the sequence representation.
A multi-label classification head predicts the utility of exposing each MQM
major category to the downstream span evaluator:
\begin{equation}
  p_c = \sigma(w_c^\top h(x,y) + b_c),
\end{equation}
where $c$ denotes an MQM major category.

The router uses multi-label supervision because multiple major-category rubrics
may be useful for the same translation instance; the label-construction details
are provided in Appendix~\ref{appendix:router}.

During inference, the router keeps the highest-scoring category and optionally
activates additional categories according to score thresholds and relative
confidence margins. The routing objective is recall-oriented activation rather
than recovery of a single gold MQM category.

\paragraph{Cascaded rubric routing.}
After category selection, the method dynamically determines how much subtype
exploration is needed. A correctness gate first predicts whether the current
translation can be accepted directly:
\begin{equation}
  p_{\mathrm{ok}}
  =
  \sigma(w_{\mathrm{ok}}^\top h(x,y)+b_{\mathrm{ok}}).
\end{equation}

If the correctness gate does not stop the item, the downstream span evaluator first runs with the Compact subtype budget. An additional expansion gate then
predicts whether broader rubric exploration is necessary:
\begin{equation}
  p_{\mathrm{exp}}
  =
  \sigma(w_{\mathrm{exp}}^\top [h(x,y);e(x,y)] + b_{\mathrm{exp}}),
\end{equation}
where $e(x,y)$ denotes compact-evaluation features.

For items selected for expansion, the subtype-budget selector chooses the Expanded or Comprehensive budget. This cascaded design keeps the default inference path
compact while allowing more complex translations to access broader rubric
coverage.

\subsection{Dynamic MQM-based Span Detection}
\label{sec:method-evaluation}

Given the activated MQM categories and the selected rubric granularity, we
retrieve the corresponding subtype descriptions and compose a case-specific
evaluation rubric:
\begin{equation}
\begin{aligned}
  \mathcal{R}'(x,y)
  &= \bigl\{r \in \mathcal{R}_{\mathrm{MQM}} :
       \mathrm{Major}(r) \in \hat{\mathcal{C}}(x,y)\bigr\},\\
  \hat{\mathcal{R}}(x,y)
  &= \mathrm{Top}\text{-}\hat{k}\bigl(\mathcal{R}'(x,y)\bigr).
\end{aligned}
\end{equation}

The composed rubric is inserted into the evaluation prompt together with the
source sentence and translation. The downstream evaluator then performs
span-level error detection under the activated MQM rubric context:
\begin{equation}
    \hat{\mathcal{S}}
    =
    f_{\mathrm{eval}}(x,y,\hat{\mathcal{R}}(x,y)).
\end{equation}

Finally, overlapping spans are merged and invalid predictions are filtered
through lightweight post-processing. The overall framework dynamically exposes
different MQM evaluation spaces to different translation cases while preserving
structured MQM-based evaluation behavior.
\section{Experiments}
\label{sec:experiments}

\subsection{Experimental Setup}
\label{sec:exp-setup}

\paragraph{Datasets.}
Experiments are conducted on the 2023 Conference on Machine Translation
(WMT23) span-level QE benchmark for Chinese--English (Zh-En) and
English--German (En-De) \cite{blain2023findings}. The WMT23 development data are partitioned into
a router-training split and a held-out development split for checkpoint and
hyperparameter selection; the test sets are reserved for final evaluation. We
mainly evaluate token-level span detection. MQM
types and subtypes are used for rubric construction and prompting, but not as
direct evaluation targets.

\paragraph{Models, training, and inference.}
We use Qwen3-4B and Qwen3-8B \cite{yang2025qwen3} in thinking mode with the
vLLM backend \cite{kwon2023efficient}, and run all experiments on NVIDIA RTX
A6000 graphics processing units (GPUs). The routing modules are lightweight
classifiers based on low-rank adaptation (LoRA) \citep{hu2022lora} that
predict the MQM search space from the source sentence and translation. The
major-category router is trained on a frozen Qwen3-8B backbone with a linear
multi-label head and LoRA adapters ($r=4$, $\alpha=8$, dropout $0.05$) for
$3$ epochs, while the rubric-granularity router uses a slightly larger LoRA
configuration ($r=8$, $\alpha=16$, dropout $0.05$) and is trained for $6$
epochs. Both routers use the 16-bit brain floating-point format (bf16) for
training. For span evaluation, we use
deterministic decoding with \texttt{temperature}=0.0, \texttt{top\_p}=1.0,
and no sampling. Unless otherwise specified, vLLM inference uses maximum
generation length $3072$, maximum model length $8192$, GPU memory utilization
$0.90$, and batch size $8$ per GPU.

\paragraph{Evaluation metrics.}
We report word-level precision (P), recall (R), F1, and MCC. Predicted spans are
converted into word-level labels and compared against gold error-span labels.

\subsection{Baselines}
\label{sec:baselines}

We compare against baselines with different MQM rubric exposure strategies.

\paragraph{Direct prediction.}
The model directly predicts erroneous spans without a fine-grained MQM subtype
rubric.

\paragraph{Static MQM rubric.}
All inputs are evaluated using the same predefined MQM rubric space. This
baseline uses rubric-guided evaluation without instance-specific adaptation.

\paragraph{Zero-shot self-routing.}
Using Qwen3-8B, the evaluator first predicts relevant MQM error types and then
reuses their definitions as a sample-specific rubric for a second
span-detection pass, without training a separate router. The full two-pass
prompts and inference procedure are provided in
Appendix~\ref{app:zero-shot-self-routing}.

\subsection{Experimental Variants}
\label{sec:exp-routing}

We compare three routing variants. \textit{Major-category selection} exposes criteria only from the selected MQM major categories. \textit{Fixed subtype-budget selection} combines the selected major categories with one fixed subtype budget. \textit{Dynamic subtype-budget selection} begins with the Compact budget and selects Expanded or Comprehensive only for items that pass the expansion gate.

\subsection{Main Results}
\label{sec:main-results}

Table~\ref{tab:main-results} reports the word-level QE results on WMT23 Zh-En
and En-De benchmarks.

\paragraph{Overall performance.}
The proposed framework consistently outperforms both direct Qwen3 baselines and
previous QE systems across both language pairs. On Zh-En, Ours-8B achieves the
best performance with 33.78 F1 and 32.40 MCC, substantially improving over
DCSQE and CometKiwi. On En-De, Ours-8B also obtains the highest F1 and MCC,
outperforming DCSQE by 2.33 F1 points and 2.46 MCC points. These results show that instance-specific construction of the evaluation search space improves span-level error localization over direct prompting in the evaluated settings.

\paragraph{Comparison with LLM baselines and recall improvements.}
Compared with vanilla Qwen3 models of the same scale, the proposed framework
consistently yields large MCC improvements across both language pairs. On
Zh-En, Ours-4B and Ours-8B improve over their corresponding Qwen3 baselines by
9.55 and 9.04 MCC points, respectively. Similar trends are also observed on
En-De. In addition, the proposed method achieves substantially higher recall
while maintaining competitive precision. On Zh-En, both Ours-4B and Ours-8B
reach recall values close to 70\%, compared with much lower recall from
vanilla Qwen3 baselines. On En-De, Ours-8B further improves recall to 60.16\%,
leading to the overall best F1 and MCC results.

Zero-shot self-routing underperforms the learned router, indicating that
reliable case-specific rubric allocation benefits from lightweight
supervision. Specific prompts and settings for this zero-shot variant are
provided in Appendix~\ref{app:zero-shot-self-routing}.

\begin{table*}[t]
\centering
\small
\renewcommand{\arraystretch}{1.18}
\setlength{\tabcolsep}{7pt}

\begin{tabular}{lcccccccc}
\hline
& \multicolumn{4}{c}{WMT23 Zh-En} & \multicolumn{4}{c}{WMT23 En-De} \\
\cline{2-5} \cline{6-9}

\textbf{Method}
& \textbf{P} & \textbf{R} & \textbf{F1} & \textbf{MCC}
& \textbf{P} & \textbf{R} & \textbf{F1} & \textbf{MCC} \\
\hline

CometKiwi
& -- & -- & -- & 26.90
& -- & -- & -- & 21.50 \\

DCSQE
& -- & -- & 28.61 & 28.12
& -- & -- & 30.61 & 27.11 \\

GEMBA
& -- & -- & 18.21 & 16.11
& -- & -- & 18.17 & 13.21 \\
\hline

Qwen3-4B
& 17.04 & 45.79 & 24.84 & 19.70
& 16.31 & 16.19 & 16.25 & 10.23 \\

Qwen3-8B
& 18.22 & 53.55 & 27.18 & 23.36
& 19.16 & 23.43 & 21.08 & 14.76 \\

Qwen3-14B
& 21.94 & 44.78 & 29.45 & 24.52
& 23.99 & 20.83 & 22.30 & 17.16 \\
\hline

Zero-shot (8B)
& 21.04 & 51.56 & 29.88 & 25.70
& 18.49 & 24.02 & 20.90 & 14.43 \\
\hline

Ours-4B
& 19.71 & 69.51 & 30.71 & 29.25
& \textbf{31.75} & 31.40 & 31.57 & 26.60 \\

Ours-8B
& \textbf{22.29} & \textbf{69.80} & \textbf{33.78} & \textbf{32.40}
& 22.68 & \textbf{60.16} & \textbf{32.94} & \textbf{29.57} \\
\hline

\end{tabular}

\caption{
Word-level QE results on WMT23 Zh-En and En-De.
All values are reported as percentages.
Results of CometKiwi, DCSQE, and GEMBA in this table are taken from \citet{geng2025alleviating}.
}
\label{tab:main-results}

\end{table*}

\subsection{Ablation on Dynamic Rubric Routing}
\label{sec:ablation-routing}

We further conduct ablation studies to analyze the contribution of different routing components.
Table~\ref{tab:ablation-routing} reports results on the Zh-En WMT23 benchmark using Qwen3-8B.
Complete ablation results for all settings are provided in the appendix.

\begin{table}[t]
\centering
\small
\renewcommand{\arraystretch}{1.18}
\setlength{\tabcolsep}{5pt}
\begin{tabular}{lcccc}
\hline
Method & P & R & F1 & MCC \\
\hline
Compact & 14.80 & 65.63 & 24.16 & 21.48 \\
Expanded & 15.40 & 65.99 & 24.98 & 22.48 \\
Comprehensive & 10.89 & 76.50 & 19.07 & 16.80 \\
+ Correct Gate & 13.70 & 70.45 & 22.95 & 20.74 \\
+ Expansion Gate & 18.93 & 71.79 & 29.96 & 28.88 \\
\hline
Full Dynamic Routing & \textbf{22.29} & \textbf{69.80} & \textbf{33.78} & \textbf{32.40} \\
\hline
\end{tabular}
\caption{
Ablation results of dynamic rubric routing on Zh-En WMT23 using Qwen3-8B.
Compact and Comprehensive correspond to fixed rubric configurations, while the remaining variants progressively introduce adaptive routing components.
}
\label{tab:ablation-routing}
\end{table}

The results reveal a clear trade-off between compact and exhaustive rubric
exploration. Fully comprehensive rubric spaces substantially improve recall,
but also introduce many false positives, leading to lower precision and MCC.
In contrast, compact rubric configurations maintain better precision but fail
to sufficiently cover complex error cases.

The Correct Gate partially alleviates this issue by filtering likely
error-free samples before rubric expansion, while the Expansion Gate further
improves precision by selectively enlarging the evaluation search space only
for difficult cases. These results support the core assumption of this work:
different translation instances require different MQM evaluation search spaces,
and dynamic rubric routing enables more effective allocation of MQM subtype
exploration.

\section{Analysis }
\subsection{Router Reliability Analysis}
\label{sec:analysis-router-reliability}

Before analyzing the final span-level predictions, we first examine whether the
routing modules can provide a reliable evaluation search space. The proposed
framework relies on two lightweight decisions before fine-grained span
detection: the major-category router, which selects a small set of MQM major
categories, and the correctness gate, which filters translations that are likely
to contain no annotated error. These two components are designed to reduce
unnecessary rubric exposure while preserving recall for error-bearing cases.

\paragraph{Major-category coverage.}
We first evaluate how many predicted MQM major categories are needed to cover the major categories corresponding to effective errors that can be identified during multi-category reasoning. Table~\ref{tab:major-router-coverage} reports the cumulative recall when at most one, two, or three major categories are activated. On Zh-En, one major category already covers 87\% of effective
error cases, while allowing up to two categories increases the coverage to
97\%. With at most three activated categories, the coverage further reaches
99\%. On En-De, the distribution is more dispersed: one category covers 69\%,
two categories cover 81\%, and three categories cover 95\% of effective error
cases.

\begin{table}[t]
\centering
\small
\begin{tabular}{lccc}
\hline
\textbf{Language Pair} & \textbf{$\leq$1} & \textbf{$\leq$2} & \textbf{$\leq$3} \\
\hline
Zh-En & 0.87 & 0.97 & 0.99 \\
En-De & 0.69 & 0.81 & 0.95 \\
\hline
\end{tabular}
\caption{Cumulative recall of effective errors covered by different numbers of activated MQM major categories.}
\label{tab:major-router-coverage}
\end{table}

These results show that \textbf{most effective errors can be covered by a small number
of major categories}. Therefore, we use at most three predicted major categories
in the final system. This setting \textbf{preserves high category-level recall while
avoiding the noisy behavior caused by activating the full MQM category space for
every instance}.

\paragraph{Correctness gate.}
The correctness gate is deliberately conservative: its primary objective is to
avoid discarding translations that still contain errors, rather than to
maximize overall binary accuracy. Table~\ref{tab:correctness-gate} reports the
confusion matrices at the selected threshold $\tau_{\mathrm{ok}}=0.65$.
Here, G and P denote gold and predicted labels, while OK and Bad denote
error-free and error-bearing cases, respectively.

\begin{table}[t]
\centering
\small
\setlength{\tabcolsep}{4pt}
\begin{tabular}{lcccc}
\hline
\textbf{Dataset} &
\shortstack{\textbf{G-OK}\\$\to$ \textbf{P-OK}} &
\shortstack{\textbf{G-OK}\\$\to$ \textbf{P-Bad}} &
\shortstack{\textbf{G-Bad}\\$\to$ \textbf{P-OK}} &
\shortstack{\textbf{G-Bad}\\$\to$ \textbf{P-Bad}} \\
\hline
Zh--En & 350 & 759 & 36 & 519 \\
En--De & 507 & 624 & 38 & 718 \\
\hline
\end{tabular}
\caption{Confusion matrices of the correctness gate. G and P denote gold and
predicted labels, respectively.}
\label{tab:correctness-gate}
\end{table}

Overall accuracy is 52.2\% on Zh--En and 64.9\% on En--De, but error recall is
93.51\% and 94.97\%, yielding false-discard rates of only 6.49\% and 5.03\%.
The gate filters 23.20\% of all Zh--En examples and 28.88\% of En--De examples
while retaining approximately 95\% of erroneous translations. Among examples
predicted as error-free, 90.67\% on Zh--En and 93.03\% on En--De are truly
Gold-OK. A threshold sweep shows the expected trade-off: increasing the
threshold reduces false discards but also filters fewer examples. We select
$\tau_{\mathrm{ok}}=0.65$ on the development set as a practical balance between
safe filtering and computational savings.

\subsection{Dynamic Rubric Complexity Analysis}
\label{sec:analysis-complexity}

\begin{figure}[t]
    \centering
    \includegraphics[width=\linewidth]{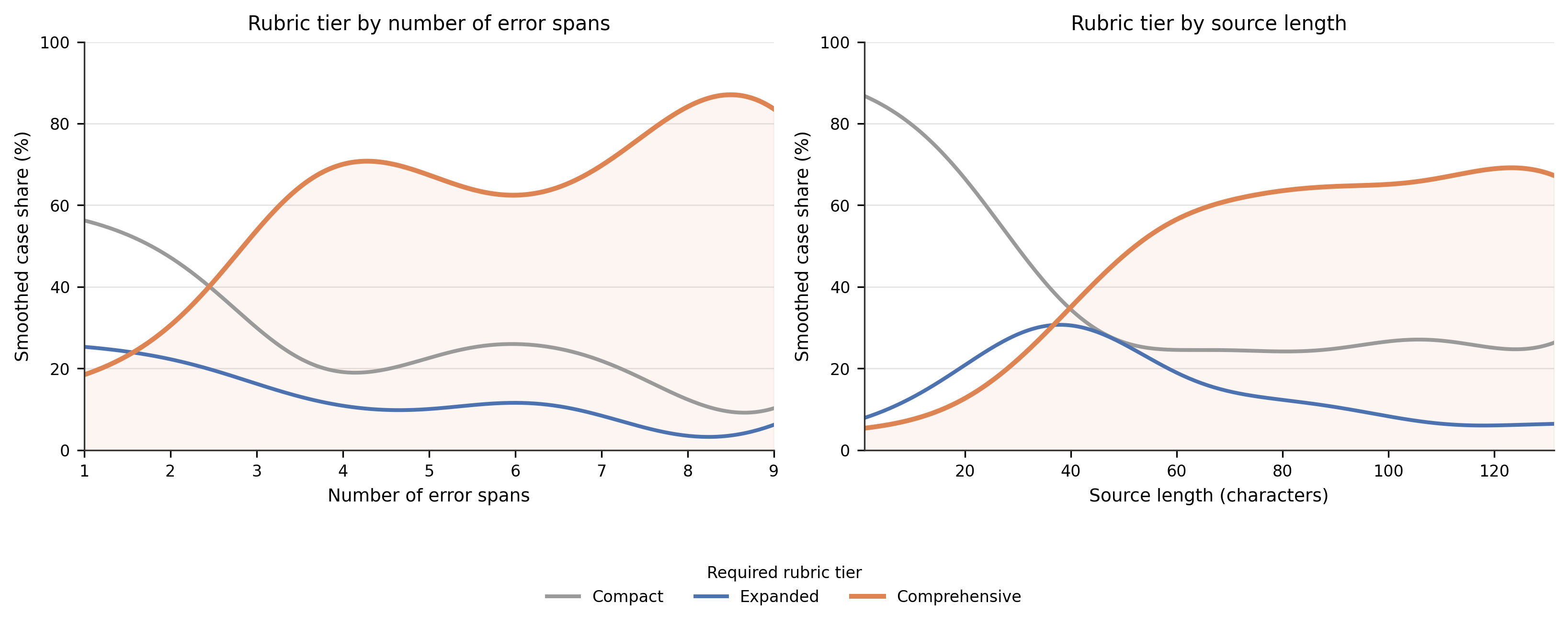}
    \caption{
    Smoothed distribution of required rubric tiers under different translation complexities.
    Left: trend with respect to the number of error spans.
    Right: trend with respect to source sentence length.
    }
    \label{fig:rubric_trend}
\end{figure}

\begin{figure}[t]
    \centering
    \includegraphics[width=\linewidth]{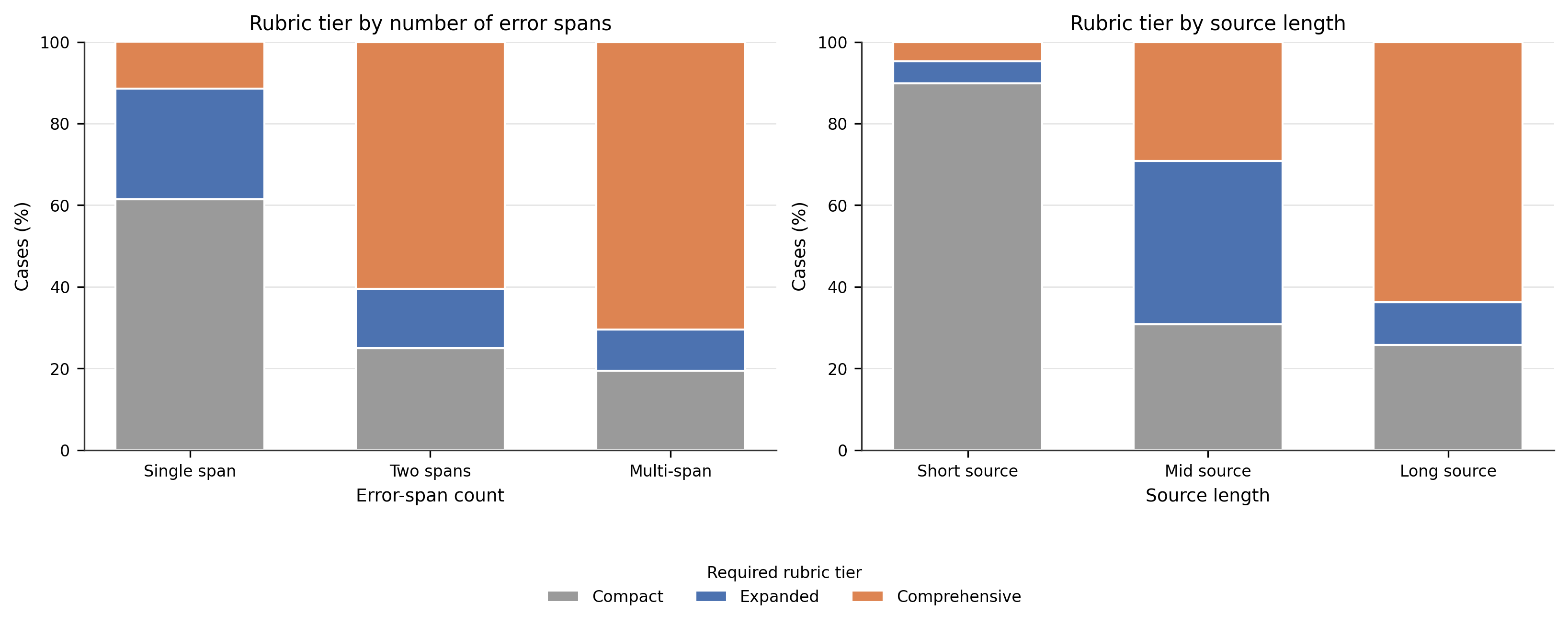}
    \caption{
    Bucketed statistics of required rubric tiers.
    Left: grouped by error-span count.
    Right: grouped by source sentence length.
    }
    \label{fig:rubric_bucket}
\end{figure}

To better understand why dynamic rubric routing is necessary, we analyze how the required rubric granularity changes with translation complexity.
Figure~\ref{fig:rubric_trend} and Figure~\ref{fig:rubric_bucket} present the distribution of required rubric tiers on the Zh-En WMT23 benchmark using Qwen3-8B. Results for other settings are provided in the appendix.

We divide the evaluation search space into three rubric tiers:
\textit{Compact}, \textit{Expanded}, and \textit{Comprehensive}.
Compact rubrics contain a small number of highly relevant MQM subtypes, while
Comprehensive rubrics expose a substantially larger fine-grained search space.

Figure~\ref{fig:rubric_trend} shows a \textbf{clear correlation between translation complexity and required rubric granularity}.
As the number of error spans increases, the proportion of cases requiring Comprehensive rubric exploration rises steadily, while Compact configurations become less effective.
A similar trend can be observed with source sentence length:
longer and semantically denser inputs increasingly require larger evaluation search spaces.

Figure~\ref{fig:rubric_bucket} further illustrates this phenomenon using coarse-grained buckets.
For single-span cases and short inputs, Compact rubric settings dominate the distribution.
In contrast, multi-span and long-source cases show a much larger proportion of Comprehensive routing decisions.
The Expanded tier mainly appears in medium-complexity regions, acting as a transition between compact and exhaustive exploration.

These results support the central motivation of our framework:
\textbf{different translation instances require substantially different evaluation search spaces}.
Instead of applying a uniform rubric configuration to all samples, the router learns to adaptively allocate rubric granularity according to the estimated complexity of the current case.
This dynamic allocation enables the evaluator to preserve recall on difficult samples while avoiding unnecessary search expansion on simpler cases.

\subsection{Case Study}
\label{sec:case-study}
\begin{table*}[!t]
\footnotesize
\centering
\renewcommand{\arraystretch}{1.15}
\setlength{\tabcolsep}{4pt}

\begin{tabular}{p{1.65cm} p{9.7cm} c c c}
\hline

\textbf{Setting} & \textbf{Detected Spans} & \textbf{P} & \textbf{R} & \textbf{F1} \\
\hline

\textbf{Source}
&
由于中西方文化背景等原因,语言之间或多或少会存在差异,如果有机会能看到作者原版相信会收获更多。
& - & - & - \\
\hline

\textbf{Target with Gold Spans}
&
Due to the cultural context \textcolor{red}{of the Midwest}, there are more or less differences between the languages, and \textcolor{red}{the author believes} that if \textcolor{red}{he} gets a chance to see the original, \textcolor{red}{he} will reap more.
& - & - & - \\
\hline

\textbf{Baseline}
&
Due to the cultural context of the \textcolor{red}{Midwest}, there are more or less differences between the languages, and the author believes that if he gets a chance to see the original, he will reap more.
& 1.000 & 0.111 & 0.200 \\
\hline

\textbf{Compact Rubric}
&
Due to the cultural context of the \textcolor{red}{Midwest}, there are more or less differences between the languages, and the author believes that if \textcolor{red}{he} gets a chance to see the original, he will reap more.
& 1.000 & 0.222 & 0.364 \\
\hline

\textbf{Expanded Rubric}
&
Due to the cultural context of \textcolor{red}{the Midwest}, there are more or less differences between the languages, and \textcolor{red}{the author believes} that if \textcolor{red}{he gets} a chance to see the original, \textcolor{red}{he} will reap more.
& 0.875 & 0.778 & \textbf{0.824} \\
\hline

\textbf{Comprehensive Rubric}
&
Due to \textcolor{red}{the cultural context of the Midwest}, there are \textcolor{red}{more or less differences between the languages}, and \textcolor{red}{the author believes} that \textcolor{red}{if he gets a chance to see the original}, \textcolor{red}{he will reap more}.
& 0.310 & 1.000 & 0.474 \\
\hline

\end{tabular}

\caption{Case study under different rubric granularities.}
\label{tab:case-study}

\end{table*}

Table~\ref{tab:case-study} presents one example under different subtype budgets. Gold and predicted error spans are highlighted in red.

As shown in Table~\ref{tab:case-study}, different rubric granularities substantially influence the span exploration behavior of the evaluator. The baseline setting is relatively conservative and only detects the core mistranslation span, resulting in low recall. As the rubric search space expands, the model gradually identifies more semantically related error spans. The medium-rubric configuration achieves the best overall performance by covering the major error structures while maintaining relatively compact span boundaries. In contrast, the full-rubric setting further increases recall but also introduces substantially noisier spans and boundary errors.

\section{Conclusion}

This paper studies the role of MQM rubrics in LLM-based span-level translation quality evaluation and argues that rubrics function not only as explicit evaluation criteria, but also as evaluation search spaces that strongly influence the exploration behavior of large language models during error localization. Experimental analysis shows a clear relationship between rubric granularity and evaluation behavior: compact rubric spaces often improve precision but suppress useful exploration and reduce recall, while excessively large rubric spaces encourage broader error exploration at the cost of substantial false positives.

Based on this observation, we propose a case-specific dynamic rubric framework that dynamically allocates MQM evaluation granularity according to the complexity of individual translation instances. Instead of applying a fixed static rubric configuration to all samples, the proposed framework combines major-category routing, cascaded rubric expansion, and adaptive subtype selection to construct instance-level evaluation search spaces for span-level QE.

Experiments on WMT23 Zh-En and En-De benchmarks demonstrate that dynamic rubric allocation achieves substantially better overall evaluation performance than direct prediction and static MQM prompting. Further analysis reveals that different translation instances exhibit substantially different rubric granularity preferences, suggesting that evaluation variability in LLM-based QE is influenced not only by model capability itself, but also by how evaluation spaces are exposed to the evaluator. Dynamic rubric allocation provides a relatively lightweight mechanism for adapting to different data distributions and annotation preferences, enabling LLM evaluators to better express their underlying error detection capabilities under diverse evaluation conditions.

\section{Limitations}

Several limitations remain in the current work. First, the proposed framework still relies on supervised routing modules and F1-based granularity signals derived from training gold spans. Although the method shows consistent improvements on WMT23 benchmarks, its robustness under different annotation standards, evaluation preferences, and unseen datasets remains insufficiently explored.

Second, the current study mainly focuses on MQM-based span-level QE with Qwen-based models and a limited set of language directions. It remains unclear whether similar dynamic evaluation-space allocation strategies generalize to other LLM evaluators, multilingual settings, or broader structured evaluation tasks beyond translation quality estimation.

In addition, while the experiments suggest that evaluation-space exposure strongly affects exploration behavior during span detection, the interaction between rubric exposure, reasoning dynamics, and span generation is still largely empirical. The current framework dynamically adjusts rubric granularity, but does not explicitly model the underlying exploration process of LLM evaluators. Future work may further investigate exploration-aware decoding, adaptive evaluation-space control, and more general dynamic rubric mechanisms for structured LLM evaluation.

\section*{Acknowledgments}

We would like to thank the anonymous reviewers for their insightful comments.
Shujian Huang is the corresponding author. This work is supported by the
National Science Foundation of China (No.~62376116), the Fundamental Research
Funds for the Central Universities (No.~2024300507), and the Fundamental and
Interdisciplinary Disciplines Breakthrough Plan of the Ministry of Education of
China (No.~JYB2025XDXM118).

\bibliography{custom}

\appendix

\section{MQM Hierarchy and Prompt Templates}
\label{app:mqm-hierarchy-prompts}

This appendix documents the MQM hierarchy used by the pipeline and the prompt
templates used in the main inference variants. The templates below are shown
before model-specific chat-template tokens are inserted. In the implementation,
they are rendered with \texttt{tokenizer.apply\_chat\_template}; when thinking
mode is enabled, the rendered prompt is additionally primed with
\verb|<think>Okay,|.

\subsection{MQM Hierarchy}
\label{app:mqm-hierarchy}

The full MQM taxonomy is organized as a
tree. For this work, we use seven broad MQM major types as the top-level routing
space. Table~\ref{tab:mqm-hierarchy} shows representative flattened subtypes
under each major type.

\begin{table*}[t]
\centering
\small
\setlength{\tabcolsep}{6pt}
\renewcommand{\arraystretch}{1.12}
\begin{tabular}{p{0.18\textwidth}p{0.74\textwidth}}
\toprule
\textbf{Major type} & \textbf{Representative flattened subtypes} \\
\midrule
Accuracy & Mistranslation; Number; Date/time; Overtranslation; Omission \\
Audience appropriateness & Locale-specific content; Language-dependent logic; Offensive \\
Design and markup & Font; Kerning; Layout; Truncation/text expansion; Markup tag \\
Linguistic conventions & Grammar; Agreement; Word order; Punctuation; Spelling \\
Locale conventions & Date format; Currency format; Quote mark type; Shortcut key \\
Style & Language register; Awkward style; Unidiomatic style; Inconsistent style \\
Terminology & Wrong term; Inconsistent use of terminology; Organization terminology \\
\bottomrule
\end{tabular}
\caption{Representative MQM rubric units used in the top-level routing space.
The displayed major types and flattened subtypes are illustrative examples;
the complete MQM hierarchy contains additional categories and subcategories.}
\label{tab:mqm-hierarchy}
\end{table*}

All lower-level nodes under each MQM major category are flattened into subtype
candidates, where each subtype is associated with a short description from the
original MQM hierarchy. For example, the Accuracy category contains subtypes
such as Mistranslation and Number, while Linguistic conventions contains
subtypes such as Grammar and Agreement. This flattened representation preserves
the hierarchical semantic information while simplifying the inference interface.

In the proposed framework, the rubric is constructed by combining the
model-predicted MQM subtype names with their corresponding descriptions. The
model first predicts relevant MQM major categories for a given
source--translation pair, and the selected categories are then expanded into
description-enhanced subtype rubric units for span-level error detection. Some MQM
categories, such as Locale conventions, contain multiple parent-linked entries
including date format, currency format, and shortcut key conventions, which are
also treated as flattened subtype candidates.

The hierarchy also provides concrete guidance for interpreting subtype
behaviors. For instance, Number errors correspond to incorrect numeric values
in translation, Overtranslation refers to targets that become unnecessarily
more specific than the source, and Wrong term captures domain-specific
terminology mismatches. Locale-specific content covers cases where the
translation becomes inappropriate under the target locale, while Unidiomatic
style describes grammatically correct but unnatural expressions. In addition,
Truncation/text expansion refers to translations whose length no longer fits
the intended layout or formatting constraints.

\subsection{Baseline span prompts}

All span-annotation prompts below require JavaScript Object Notation (JSON)
output. Severity is predicted directly by the downstream evaluator together
with the error span and category; it is not assigned by a routing module or a
post-hoc heuristic. The output vocabulary is restricted to \texttt{minor},
\texttt{major}, and \texttt{critical}.

\begin{promptbox}{Baseline span prompt}
SYSTEM:
You are an MQM annotation assistant. Output JSON only, no explanation.
Required output schema: {"items":[{"error_text":"...","category":"...","severity":"minor|major|critical"}]}
Rules:
1) error_text must be an exact text span copied from TARGET.
2) Do not output indexes; output only the erroneous text content.
3) category must be a string, and severity must be exactly one of "minor", "major", or "critical".
4) If no error, output {"items":[]}.
5) Never output extra keys or text.

USER:
Task: find erroneous text in TARGET and classify each with category + severity.
Return each error as exact copied text from TARGET.

SOURCE:
{source}

TARGET:
{target}
\end{promptbox}

\subsection{Zero-shot Self-routing Prompts and Settings}
\label{app:zero-shot-self-routing}

The zero-shot self-routing baseline uses Qwen3-8B with the \texttt{predicted}
route of the \texttt{mqm\_rubric\_eval} pipeline. It is a two-pass,
self-conditioned rubric procedure with no learned routing module. In the first
pass, Qwen3-8B compares the source and target and predicts zero or more error
types from a predefined MQM label inventory. The predicted labels and their
definitions are then converted into a sample-level rubric. In the second pass,
the same Qwen3-8B evaluator receives this self-predicted rubric and identifies
exact target spans, categories, and severity levels. Unless noted otherwise,
this baseline uses the same inference settings as those in
Section~\ref{sec:exp-setup}.

The first pass uses the following prompt. The
\texttt{allowed\_labels\_with\_descriptions} field is rendered as one label and
its definition per line; the inventory contains Accuracy/Addition,
Accuracy/Creative Reinterpretation, Accuracy/Mistranslation,
Accuracy/Omission, Accuracy/Source language fragment, Fluency/Character
encoding, Fluency/Grammar, Fluency/Inconsistency, Fluency/Punctuation,
Fluency/Register, Fluency/Spelling, Non-translation!, Style/Bad sentence
structure, Style/Unnatural or awkward, Terminology/Inappropriate for context,
and Terminology/Inconsistent.

\begin{promptbox}{Zero-shot self-routing: error-type prediction prompt}
SYSTEM:
You are an MQM error-type classifier for translation quality evaluation. Output JSON only.

Return exactly one JSON object with the schema:
{"predicted_error_types":[str], "rationale":str, "confidence":number|null}

Task:
Identify translation errors in TARGET compared with SOURCE and assign labels only from ALLOWED_LABELS.

Guidelines:
- Use SOURCE and TARGET together to detect translation problems.
- Select all labels that clearly apply.
- Do not invent labels outside ALLOWED_LABELS.
- If multiple labels apply, include all of them.
- Do not avoid labeling simply because the exact label wording is imperfect; choose the closest applicable label.
- Use [] only when the translation contains no identifiable error from the allowed dictionary.

Decision principle:
- Prefer identifying clear errors over returning an empty list.
- Only return [] when you are confident that the translation contains no identifiable error from the allowed dictionary.

Output rules:
- predicted_error_types must be a JSON array of labels from ALLOWED_LABELS.
- rationale must be a short explanation grounded in the translation.
- confidence must be a number from 0 to 1, or null.
- Do not output extra keys or extra text.

USER:
SOURCE:
{source}

TARGET:
{target}

Allowed labels:
{allowed_labels_with_descriptions}
\end{promptbox}

Each predicted type is paired with its dictionary definition and converted to
the following rubric structure; multiple predictions produce multiple entries
in \texttt{checks}.

\begin{promptbox}{Zero-shot self-routing: self-predicted rubric}
{
  "scope": "sample",
  "checks": [
    {
      "mqm_error_type_display_name": "{predicted_error_type}",
      "error_type_description": "{error_type_definition}",
      "rationale": "Prioritize checking for {predicted_error_type}."
    }
  ]
}
\end{promptbox}

If the first pass returns an empty label set, the pipeline does not assume that
the translation is error-free. Instead, it inserts the following generic
fallback check:

\begin{promptbox}{Zero-shot self-routing: empty-prediction fallback}
{
  "scope": "sample",
  "checks": [
    {
      "mqm_error_type_display_name": "GENERIC",
      "error_type_description": "Unknown error type. There may or may not be errors. Do not assume correctness; if clear errors are present, they must be identified.",
      "rationale": "Fallback rubric for empty predicted type set."
    }
  ]
}
\end{promptbox}

The second pass uses the resulting rubric to localize errors:

\begin{promptbox}{Zero-shot self-routing: rubric-conditioned span prompt}
SYSTEM:
You are an MQM annotation assistant. Output JSON only, no explanation.
Required output schema: {"items":[{"error_text":"...","category":"...","severity":"minor|major|critical"}]}
Rules:
1) error_text must be an exact text span copied from TARGET.
2) Do not output indexes; output only the erroneous text content.
3) category must be a string, and severity must be exactly one of "minor", "major", or "critical".
4) If no error, output {"items":[]}.
5) Never output extra keys or text.

USER:
Task: find erroneous text in TARGET and classify each with category + severity.
Return each error as exact copied text from TARGET.
Use rubric.checks as primary guidance for locating errors.
If no error is found, output {"items":[]}.

SOURCE:
{source}

TARGET:
{target}

RUBRIC_JSON:
{self_predicted_rubric}
\end{promptbox}

\subsection{Merge Prompts}
\label{app:merge-prompts}

The merge pipeline first uses the major-category router to predict relevant MQM
major types for each source--translation pair. For each predicted major type,
all corresponding subtype candidates under that subtree are activated and used
for span-level evaluation. The predicted spans and error types from different
major-type subtrees are then merged into the final output.

The first prompt below is used by
\texttt{format\_hierarchy\_subtree\_predict\_prompt}. In the standard merge
variant, the minimal-span rule is disabled; in the minimal-span variant, the
additional minimal-span line is inserted.

\begin{promptbox}{Hierarchy merge: major-subtree detection prompt}
SYSTEM:
You are an MQM translation quality evaluation assistant. Output JSON only.

Return exactly one JSON object with the schema:
{"items":[{"error_text":"...","error_type":"...","error_type_description":"...","severity":"minor|major|critical"}]}

Task:
Evaluate the translation quality of TARGET against SOURCE.
Find translation errors in TARGET that belong to the provided MQM subtree.
If such errors exist, output one item for each detected error.

Rules:
- error_text must be exact copied text from TARGET.
- [optional for minimal-span runs] error_text must be the smallest TARGET span that still fully identifies the error; do not include surrounding correct text.
- error_type must be chosen exactly from CANDIDATE_CHECKS.error_type.
- error_type_description must exactly match the chosen candidate description.
- severity must be exactly one of "minor", "major", or "critical".
- If no matching error exists in this subtree, output {"items":[]}.
- Do not output extra keys or extra text.

USER:
MQM_TOP_LEVEL:
{top_level_name}

Instruction:
Compare SOURCE and TARGET and identify actual translation errors only.
Do not invent issues. But if a clear error exists in this MQM subtree, report it.

SOURCE:
{source}

TARGET:
{target}

CANDIDATE_CHECKS:
{candidate_checks_json}
\end{promptbox}

After the subtree outputs are merged, the pipeline can run a second span
inference pass guided by the merged predicted error types. This prompt is used by
\texttt{format\_eval\_prompt\_with\_error\_type\_\allowbreak rubric}.

\begin{promptbox}{Hierarchy merge: merged-rubric span prompt}
SYSTEM:
You are an MQM annotation assistant. Output JSON only, no explanation.
Required output schema: {"items":[{"error_text":"...","category":"...","severity":"minor|major|critical"}]}
Rules:
1) error_text must be an exact text span copied from TARGET.
2) Do not output indexes; output only the erroneous text content.
3) category must be a string, and severity must be exactly one of "minor", "major", or "critical".
4) If no error, output {"items":[]}.
5) Never output extra keys or text.

USER:
Task: identify translation errors in TARGET based on SOURCE.
Task: find erroneous text in TARGET and classify each with category + severity.
Return each error as exact copied text from TARGET.
Use rubric.checks as primary guidance for locating errors.
If no error is found, output {"items":[]}.

SOURCE:
{source}

TARGET:
{target}

RUBRIC_JSON:
{
  "scope": "sample",
  "checks": [
    {
      "mqm_error_type_display_name": "{predicted_mqm_type}",
      "error_type_description": "{description}",
      "rationale": "Use this predicted MQM type as guidance: {predicted_mqm_type}."
    }
  ]
}
\end{promptbox}

\subsection{Candidate-granularity Span Prompt}
\label{app:topk-best-prompt}

This prompt produces span outputs under candidate rubric granularities when
constructing router supervision on the training split and when conducting
post-hoc oracle analysis. It is not a test-time output-selection procedure. At
test time, the learned router selects the rubric granularity before span
evaluation using only the source--translation pair and model predictions.

\begin{promptbox}{Candidate-granularity minimal-span subtree prompt}
SYSTEM:
You are an MQM translation quality evaluation assistant. Output JSON only.

Return exactly one JSON object with the schema:
{"items":[{"error_text":"...","error_type":"...","error_type_description":"...","severity":"minor|major|critical"}]}

Task:
Evaluate the translation quality of TARGET against SOURCE.
Find translation errors in TARGET that belong to the provided MQM subtree.
If such errors exist, output one item for each detected error.

Rules:
- error_text must be exact copied text from TARGET.
- error_text must be the smallest TARGET span that still fully identifies the error; do not include surrounding correct text.
- error_type must be chosen exactly from CANDIDATE_CHECKS.error_type.
- error_type_description must exactly match the chosen candidate description.
- severity must be exactly one of "minor", "major", or "critical".
- If no matching error exists in this subtree, output {"items":[]}.
- Do not output extra keys or extra text.

USER:
MQM_TOP_LEVEL:
{major_type}

Instruction:
Compare SOURCE and TARGET and identify actual translation errors only.
Do not invent issues. But if a clear error exists in this MQM subtree, report it.

SOURCE:
{source}

TARGET:
{target}

CANDIDATE_CHECKS:
{topk_candidate_checks_json}
\end{promptbox}
\subsection{Subtype-ranking Prompt}
\label{app:subtype-ranking-prompt}

For each source--translation pair and MQM major category, the model predicts
multiple granularities of subtype candidate sets. The goal of this stage is not
to directly predict final translation errors, but to estimate which MQM subtype
definitions would be most useful as auxiliary rubric entries for later
span-level evaluation. Smaller candidate sets correspond to more compact rubric
spaces, while larger candidate sets expose broader subtype exploration spaces.

\begin{promptbox}{Subtype-ranking prompt}
SYSTEM:
You are an MQM rubric selection assistant.
Your final answer must be a JSON object with this schema:
{
  "compact":["subtype name"],
  "medium":["subtype name"],
  "large":["subtype name"],
  "full":["subtype name"]
}

Rules:
1) Select only subtype names from the provided SUBTYPE_RUBRIC_JSON.
2) Rank subtypes by usefulness as auxiliary rubric entries for judging possible translation errors in this case.
3) The selected subtypes are not final activated error labels.
4) compact should contain the most essential subtype candidates.
5) medium should expand the compact subtype space with additional useful subtype candidates.
6) large should further expand the subtype space with broader subtype coverage.
7) full should contain the most comprehensive subtype exploration space available for this major category.
8) If the subtype inventory is small, return as many subtype names as available.
9) Do not include duplicate subtype names or labels outside the rubric.

USER:
Task: select MQM subtype rubric entries that would be most useful as auxiliary rubric entries for judging possible translation errors in this SOURCE/TARGET pair.
Evaluate only the single major category shown in SUBTYPE_RUBRIC_JSON.
Do not decide the final error type. Rank subtype definitions by how helpful they are for analyzing this case.

SOURCE:
{source}

TARGET:
{target}

SUBTYPE_RUBRIC_JSON:
{
  "major_category": "{major_type}",
  "subtypes": [
    {
      "name": "{subtype_name}",
      "path": "{major_type}/{subtype_path}",
      "description": "{subtype_description}"
    }
  ]
}
\end{promptbox}

\section{Detailed Method Description}
\label{appendix:method}

\subsection{Detailed MQM Category Routing}
\label{appendix:router}

Let $\mathcal{C}$ denote the seven MQM major categories:
Accuracy, Audience appropriateness, Design and markup, Linguistic conventions,
Locale conventions, Style, and Terminology.

Given a source sentence $x$ and translation $y$, we encode the pair using a
Qwen3 causal language model with an instruction-style prompt. The router uses
the final hidden state at the last non-padding token as the sequence
representation:
\begin{equation}
  h(x,y) = H_{\ell_i}^{(L)},
  \qquad
  \ell_i = \sum_t m_{i,t} - 1,
\end{equation}
where $H^{(L)}$ denotes the last-layer hidden-state matrix and $m_{i,t}$ is
the attention mask.

A linear classification head predicts one score for each MQM category:
\begin{equation}
  z_c = w_c^\top h(x,y) + b_c,
  \qquad
  p_c = \sigma(z_c),
\end{equation}
where $c \in \mathcal{C}$.

The category supervision is constructed offline on the routing-training split.
For every training instance, we enumerate all seven categories. For each
$c\in\mathcal{C}$, the downstream evaluator performs span detection using the
rubric associated with category $c$, producing a candidate span set
$\mathcal{S}_c$. We convert both $\mathcal{S}_c$ and the training gold span set
$\mathcal{S}^{\star}$ into word-level binary labels and compute
\begin{equation}
  u_c(x,y)
  =
  F_1^{\mathrm{word}}
  \left(
    \mathcal{S}_c,
    \mathcal{S}^{\star}
  \right).
\end{equation}
The per-category scores $\{u_c(x,y):c\in\mathcal{C}\}$ measure how useful each
major-category rubric is for recovering the annotated error words. We use these
word-level F1 scores directly as soft multi-label targets:
\begin{equation}
  q_c
  =
  u_c(x,y),
  \qquad q_c\in[0,1].
\end{equation}
Thus, more than one category can receive positive supervision for the same
instance. The targets represent category-conditioned span-detection utility
rather than gold MQM error types.

The router is trained as a multi-label classifier using binary cross entropy:
\begin{equation}
\begin{aligned}
  \mathcal{L}_{\mathrm{cat}}
  = - \sum_{c \in \mathcal{C}'} \bigl[&
    \omega_c q_c \log p_c \\
    &+ (1-q_c)\log(1-p_c)
  \bigr],
\end{aligned}
\end{equation}
where $\mathcal{C}'$ optionally includes a \textsc{NoCategory} label,
$q_c$ is the utility-based soft multi-label target defined above, and
$\omega_c$ is a positive-class reweighting factor estimated from the training
split.

To reduce imbalance between error-bearing and no-error cases, all-zero MQM
examples are downsampled during training.

During inference, the router is tuned toward recall-oriented activation.
Let
\begin{equation}
c_1=\arg\max_{c\in\mathcal{C}} p_c
\end{equation}
be the highest-scoring category. Additional categories are activated if
\begin{equation}
  p_c \ge \tau_{\mathrm{cat}}
  \quad \mathrm{or} \quad
  p_c \ge p_{c_1}-\delta,
\end{equation}
up to a maximum number of activated categories.

If the optional \textsc{NoCategory} score is high while all MQM category
scores remain below a minimum confidence threshold, the router returns an
empty category set. For extremely low-confidence cases, Accuracy is used as a
fallback category because it covers common adequacy-related failures.

\subsection{Detailed Cascaded Rubric Routing}
\label{appendix:cascade}

After MQM major-category selection, the framework dynamically determines how
much subtype exploration is needed.

\paragraph{Correctness gate.}
A correctness gate first predicts whether the translation can be accepted
without additional subtype evaluation:
\begin{equation}
  a_{\mathrm{ok}}
  =
  w_{\mathrm{ok}}^\top h(x,y)+b_{\mathrm{ok}},
\end{equation}
\begin{equation}
  p_{\mathrm{ok}}
  =
  \sigma(a_{\mathrm{ok}}).
\end{equation}

The threshold $\tau_{\mathrm{ok}}=0.65$ is selected on the development set. If
$p_{\mathrm{ok}} \ge \tau_{\mathrm{ok}}$, the system directly returns a
no-error decision.

Otherwise, the evaluator proceeds with compact subtype evaluation using a
small set of highly relevant MQM subtypes.

\paragraph{Expansion gate.}
Some translations require broader rubric coverage beyond compact evaluation.
We therefore introduce an expansion gate after compact evaluation.

Let $e_3(x,y)$ denote compact-evaluation features. The expansion gate predicts
whether broader subtype exploration is necessary:
\begin{equation}
  a_{\mathrm{exp}}
  =
  w_{\mathrm{exp}}^\top [h(x,y);e_3(x,y)]
  +
  b_{\mathrm{exp}},
\end{equation}
\begin{equation}
  p_{\mathrm{exp}}
  =
  \sigma(a_{\mathrm{exp}}).
\end{equation}

If $p_{\mathrm{exp}} < \tau_{\mathrm{exp}}$, the compact evaluation result is
kept. Otherwise, a budget router determines whether to expand toward a
medium-granularity subtype pool or the full subtype pool:
\begin{equation}
  \mathcal{B}=\{\mathrm{Expanded},\mathrm{Comprehensive}\},
\end{equation}
\begin{equation}
  \hat{b}
  =
  \arg\max_{b\in\mathcal{B}}
  \pi_b(x,y,e_3).
\end{equation}

The medium setting exposes a broader but still selective subtype space,
whereas the full setting activates all available subtypes under the selected
MQM categories.

\subsection{Routing Supervision and Evaluation Protocol}
\label{appendix:training}

\paragraph{Training supervision.}
Gold annotations are used on the training split only to construct router
labels. The major-category labels $q_c$ are constructed by enumerating all seven
major categories and comparing their category-conditioned predictions with the
training gold spans using word-level F1, as described above. The correctness
target is $y_{\mathrm{ok}}=1$ only when the training annotation contains no
error span. To supervise rubric expansion, we generate candidate outputs under
each granularity and compare them with the training gold spans after generation.
We select the training-only oracle exclusively by word-level F1. Let
$\mathcal{G}$ denote the three candidate granularities:
\begin{equation}
\begin{aligned}
  \mathcal{G}
  &= \{\mathrm{Compact},\mathrm{Expanded},\\
  &\qquad \mathrm{Comprehensive}\},\\
  g^\star_{\mathrm{train}}
  &= \arg\max_{g\in\mathcal{G}} F_1^{\mathrm{word}}(g).
\end{aligned}
\end{equation}
This oracle is a supervision label, not an inference rule.

The correctness gate is trained with
\begin{equation}
  \mathcal{L}_{\mathrm{ok}}
  =
  \mathrm{BCEWithLogits}
  (
  a_{\mathrm{ok}},
  y_{\mathrm{ok}}
  ).
\end{equation}

The expansion target indicates whether the training-only oracle exceeds
Compact evaluation:
\begin{equation}
  y_{\mathrm{exp}}
  =
  \mathbb{I}
  [
  g^\star_{\mathrm{train}} \neq \mathrm{Compact}
  ],
\end{equation}
\begin{equation}
  \mathcal{L}_{\mathrm{exp}}
  =
  \mathrm{BCEWithLogits}
  (
  a_{\mathrm{exp}},
  y_{\mathrm{exp}}
  ).
\end{equation}

The budget router is trained only on examples requiring expansion, using
cross-entropy (CE):
\begin{equation}
\begin{aligned}
  \mathcal{L}_{\mathrm{bud}}
  &= \mathbb{I}[g^\star_{\mathrm{train}} \neq \mathrm{Compact}]\\
  &\quad \cdot \mathrm{CE}\!\left(
    \pi,
    \operatorname{index}_{\mathcal{B}}(g^\star_{\mathrm{train}})
  \right).
\end{aligned}
\end{equation}

The total routing objective is:
\begin{equation}
  \mathcal{L}
  =
  \mathcal{L}_{\mathrm{cat}}
  +
  \mathcal{L}_{\mathrm{ok}}
  +
  \mathcal{L}_{\mathrm{exp}}
  +
  \mathcal{L}_{\mathrm{bud}}.
\end{equation}

\paragraph{Development-set tuning.}
The held-out development set is used to select router checkpoints and fix the
routing hyperparameters: the category threshold $\tau_{\mathrm{cat}}$, relative
margin $\delta$, maximum number of activated categories, correctness threshold
$\tau_{\mathrm{ok}}=0.65$, and expansion threshold $\tau_{\mathrm{exp}}$. These
choices are frozen before test evaluation.

\paragraph{Test-time inference.}
The WMT23 test sets contain 1,664 Zh--En instances and 1,887 En--De instances.
For a test instance, the system receives only the source sentence $x$ and
translation $y$. The trained routers predict the major categories, correctness
decision, expansion decision, and rubric budget; the span evaluator then
generates the final error spans under that predicted rubric. Test annotations
and test-derived metrics are unavailable to every routing and generation step.

\paragraph{Post-hoc test analysis.}
After all test predictions have been finalized, test gold annotations are used
only to compute aggregate evaluation metrics and retrospective analyses, such
as oracle performance ceilings or granularity-preference distributions. Any
per-sample F1 or oracle-best granularity reported in an analysis is computed
after inference and is never fed back into the router or used to replace its
output. \textbf{No gold spans, per-sample F1 scores, or oracle-selected outputs
are used at test time.}

\subsection{Detailed Dynamic Span Detection}
\label{appendix:eval}

Given the selected MQM major categories
$\hat{\mathcal{C}}(x,y)$ and rubric size $\hat{k}$, the framework retrieves
the corresponding subtype descriptions from the MQM rubric space:
\begin{equation}
\begin{aligned}
    \mathcal{R}'(x,y)
    &= \bigl\{r \in \mathcal{R}_{\mathrm{MQM}} :
       \mathrm{Major}(r) \in \hat{\mathcal{C}}(x,y)\bigr\},\\
    \hat{\mathcal{R}}(x,y)
    &= \mathrm{Top}\text{-}\hat{k}\bigl(\mathcal{R}'(x,y)\bigr).
\end{aligned}
\end{equation}

The composed rubric is inserted into the evaluation prompt together with the
source sentence and translation:
\begin{equation}
    \hat{\mathcal{S}}
    =
    f_{\mathrm{eval}}
    (
    x,
    y,
    \hat{\mathcal{R}}(x,y)
    ).
\end{equation}

The evaluator is instructed to output only spans copied from the translation,
ensuring alignment with the span-level QE objective.

When multiple evaluation outputs are produced, we apply lightweight
post-processing:
\begin{itemize}
    \item highly overlapping spans are merged;
    \item duplicated predictions are removed;
    \item invalid spans not appearing in the translation are filtered;
    \item optional calibration combines think and no-think outputs.
\end{itemize}

This design allows different translation cases to access different MQM subtype
spaces while preserving structured MQM-based evaluation behavior.

\section{Ablation Settings}
\label{sec:appendix-ablation}

To better understand the contribution of each component in the proposed
framework, we conduct a series of ablation experiments with progressively
increasing levels of dynamic routing capability.

We replace the original fixed top-$k$ terminology with three semantic
search-space levels:

\begin{itemize}
    \item \textbf{Compact}: a small fine-grained rubric space;
    \item \textbf{Expanded}: a medium-sized rubric space;
    \item \textbf{Comprehensive}: the full MQM subtype space.
\end{itemize}

The evaluated settings are summarized in
Table~\ref{tab:ablation-settings}.

\begin{table*}[h]
\centering
\small
\renewcommand{\arraystretch}{1.12}
\setlength{\tabcolsep}{3pt}

\begin{tabular}{lcccc}
\hline
\textbf{Setting} &
\textbf{Correct} &
\textbf{Expand} &
\textbf{Budget} &
\textbf{Output Space} \\
\hline

Fixed Compact &
$\times$ & $\times$ & $\times$ &
Compact \\

Fixed Expanded &
$\times$ & $\times$ & $\times$ &
Expanded \\

Fixed Full &
$\times$ & $\times$ & $\times$ &
Comprehensive \\

+ Correct Gate &
$\checkmark$ & $\times$ & $\times$ &
stop / Compact \\

+ Expansion Gate &
$\checkmark$ & $\checkmark$ & $\times$ &
stop / Compact / Expanded \\

Full Dynamic &
$\checkmark$ & $\checkmark$ & $\checkmark$ &
adaptive full routing \\

\hline
\end{tabular}

\caption{Ablation settings for dynamic rubric routing.}
\label{tab:ablation-settings}

\end{table*}

The \textbf{Correct Gate} predicts whether a translation is likely to
contain errors. Samples predicted as correct terminate early
without fine-grained evaluation. Otherwise, the system enters Compact
evaluation.

The \textbf{Expansion Gate} determines whether the current Compact
search space is sufficient. If not, the evaluation space is expanded to
a larger rubric set.

Finally, the \textbf{Budget Router} controls the expansion granularity
for difficult cases, dynamically selecting between Expanded and
Comprehensive search spaces according to the estimated sample complexity.

\subsection{Ablation Results}
\label{sec:appendix-ablation-results}

Tables~\ref{tab:ablation-zhen23} and \ref{tab:ablation-ende23} present the complete ablation results on WMT23 Zh$\rightarrow$En and En$\rightarrow$De benchmarks.  We report word-level Precision (P), Recall (R), F1, and MCC.

\begin{table*}[t]
\centering
\small
\renewcommand{\arraystretch}{1.18}

\setlength{\tabcolsep}{5pt}
\begin{tabular}{llcccc}
\hline
\textbf{Model} & \textbf{Setting} & \textbf{P} & \textbf{R} & \textbf{F1} & \textbf{MCC} \\
\hline

\multirow{6}{*}{Ours-4B}
& Fixed Compact       & 13.74 & 68.53 & 22.89 & 20.39 \\
& Fixed Expanded      & 14.14 & 70.59 & 23.56 & 20.59 \\
& Fixed Comprehensive & 11.99 & 73.86 & 20.63 & 18.26 \\
& + Correct Gate      & 15.86 & 63.85 & 25.41 & 22.64 \\
& + Expansion Gate    & 17.56 & 75.71 & 28.51 & 28.04 \\
& Full Dynamic        & \textbf{19.71} & 69.51 & \textbf{30.71} & \textbf{29.25} \\
\hline

\multirow{6}{*}{Ours-8B}
& Fixed Compact       & 14.80 & 65.63 & 24.16 & 21.48 \\
& Fixed Expanded      & 15.40 & 65.99 & 24.98 & 22.48 \\
& Fixed Comprehensive & 10.89 & 76.50 & 19.07 & 16.80 \\
& + Correct Gate      & 13.70 & 70.45 & 22.95 & 20.74 \\
& + Expansion Gate    & 18.93 & 71.79 & 29.96 & 28.88 \\
& Full Dynamic        & \textbf{22.29} & 69.80 & \textbf{33.78} & \textbf{32.40} \\
\hline

\end{tabular}
\caption{Ablation results on WMT23 Zh$\rightarrow$En.}
\label{tab:ablation-zhen23}
\end{table*}

\begin{table*}[t]
\centering
\small
\renewcommand{\arraystretch}{1.18}

\setlength{\tabcolsep}{5pt}
\begin{tabular}{llcccc}
\hline
\textbf{Model} & \textbf{Setting} & \textbf{P} & \textbf{R} & \textbf{F1} & \textbf{MCC} \\
\hline

\multirow{6}{*}{Ours-4B}
& Fixed Compact       & 14.71 & 38.30 & 21.26 & 14.28 \\
& Fixed Expanded      & 13.64 & 43.59 & 20.77 & 14.52 \\
& Fixed Comprehensive & 14.09 & 42.13 & 21.11 & 14.73 \\
& + Correct Gate      & 19.66 & 36.19 & 22.46 & 16.10 \\
& + Expansion Gate    & 26.72 & 30.26 & 28.38 & 22.83 \\
& Full Dynamic        & \textbf{31.75} & 31.40 & \textbf{31.57} & \textbf{26.60} \\
\hline

\multirow{6}{*}{Ours-8B}
& Fixed Compact       & 13.00 & 45.96 & 20.27 & 13.81 \\
& Fixed Expanded      & 12.67 & 46.11 & 19.87 & 13.33 \\
& Fixed Comprehensive & 14.18 & 48.09 & 21.90 & 16.01 \\
& + Correct Gate      & 18.18 & 45.09 & 22.90 & 18.01 \\
& + Expansion Gate    & 25.16 & 30.10 & 27.41 & 21.68 \\
& Full Dynamic        & \textbf{22.68} & \textbf{60.16} & \textbf{32.94} & \textbf{29.57} \\
\hline

\end{tabular}
\caption{Ablation results on WMT23 En$\rightarrow$De.}
\label{tab:ablation-ende23}
\end{table*}

Overall, the results show that simply enlarging the evaluation search space from Compact to Comprehensive mainly improves recall, but often introduces substantial false positives and reduces precision.  The Correct Gate improves precision by filtering likely error-free samples before detailed evaluation.  The Expansion Gate further improves the balance between precision and recall by selectively enlarging the search space only for difficult cases.

Finally, the Full Dynamic configuration consistently achieves the strongest overall performance across both language directions and model scales, demonstrating the effectiveness of instance-level adaptive rubric routing.

\section{Gold-Rubric Upper-Bound Validation}
\label{sec:appendix-gold-rubric}

We conduct an additional gold-rubric experiment to examine whether more
accurate rubric allocation can improve
downstream error detection. This experiment is a post-hoc upper-bound analysis,
not a test-time variant of the proposed method. For each test instance, we
first generate outputs under the candidate rubric configurations and only then
use the reference evaluation result to identify the configuration with the best
detection performance. The evaluator itself is never provided with gold error
text, positions, or span boundaries. Thus, test gold is used only for
retrospective rubric selection after generation, consistent with the separation
described in Section~\ref{appendix:training}.

\begin{table}[t]
\centering
\small
\renewcommand{\arraystretch}{1.12}
\setlength{\tabcolsep}{5pt}
\begin{tabular}{lcccc}
\toprule
\textbf{Dataset} & \textbf{P} & \textbf{R} & \textbf{F1} & \textbf{MCC} \\
\midrule
Zh--En & 28.13 & 74.47 & 40.83 & 40.02 \\
En--De & 25.64 & 63.98 & 36.61 & 33.80 \\
\bottomrule
\end{tabular}
\caption{Word-level results of the post-hoc gold-rubric upper bound. All values
are reported as percentages.}
\label{tab:gold-rubric-results}
\end{table}

The gold-rubric upper bound substantially outperforms automatic rubric
allocation. This is expected because its rubric is selected with access to the
reference evaluation result after candidate outputs have been generated. It is
therefore not a fair comparison with the proposed test-time method. Instead,
the result shows that downstream error detection can improve further when the
evaluator receives a more suitable case-specific rubric, motivating future work
on more reliable rubric allocation.

We further measure whether the evaluator follows the supplied rubric or assigns
error categories outside it.

\begin{table}[t]
\centering
\small
\renewcommand{\arraystretch}{1.12}
\setlength{\tabcolsep}{6pt}
\begin{tabular}{lr}
\toprule
\textbf{Statistic} & \textbf{Count} \\
\midrule
Total predicted error-category assignments & 1,773 \\
Contained in the prompt rubric & 1,758 \\
Outside the prompt rubric & 15 \\
Rubric adherence rate & \textbf{99.15\%} \\
\bottomrule
\end{tabular}
\caption{Adherence of predicted error categories to the supplied rubric.}
\label{tab:gold-rubric-adherence}
\end{table}

Among 1,773 predicted error-category assignments, 1,758 are contained in the
supplied rubric and only 15 fall outside it, giving a 99.15\% adherence rate.
The evaluator therefore almost always predicts within the supplied rubric,
showing that the rubric restricts the considered error categories rather than
merely providing a weak textual bias. Together, the upper-bound improvement
and high adherence support two conclusions: a more suitable case-specific
rubric can substantially improve error detection, and the supplied rubric
effectively controls the evaluator's error-search space.

\section{Sentence-level Evaluation}
\label{sec:appendix-sentence-level}

Our primary task is fine-grained QE through error-span localization rather than
direct sentence-level quality prediction. This distinction is important because
recent QE systems may explicitly optimize both granularities. For example,
\citet{geng2025alleviating} jointly optimize sentence-level regression and
word-level sequence labeling, and combine the regression output with an MQM
score derived from predicted spans at inference time. In contrast, our method
receives neither sentence-level regression supervision nor sentence-level MQM
score supervision.

For this complementary analysis, each predicted error item contains a severity
label generated directly by the span evaluator. The model selects exactly one
of \textit{minor}, \textit{major}, or \textit{critical}; severity is not
assigned by the learned routers or inferred afterward from the predicted error
category. The primary word-level Precision, Recall, F1, and MCC metrics use the
predicted span labels and do not depend on severity; severity is used only for
the complementary sentence-level aggregation below. Given the predicted error
set $\hat{\mathcal{E}}(x,y)$, let $\hat{v}_i$ denote the model-predicted
severity of error item $e_i$. We aggregate the predicted severities into an MQM
penalty:
\begin{equation}
\begin{gathered}
  \hat{p}(x,y)
  = \sum_{e_i \in \hat{\mathcal{E}}(x,y)} w(\hat{v}_i),\\
  \bigl(w(\text{minor}),w(\text{major}),w(\text{critical})\bigr)
  = (1,5,10).
\end{gathered}
\end{equation}
Lower penalties indicate better translations. Its length-normalized form is
\begin{equation}
    \hat{p}_{\mathrm{norm}}(x,y) = \frac{\hat{p}(x,y)}{L_y},
\end{equation}
where $L_y$ is the number of tokens in the translation. Table~\ref{tab:sentence-correlation}
reports correlations between the unnormalized predicted penalty
$\hat{p}(x,y)$ and the corresponding human MQM penalty, which is the scoring
configuration used to obtain the reported values. All correlation values are
multiplied by 100.

\begin{table}[t]
\centering
\small
\renewcommand{\arraystretch}{1.12}
\setlength{\tabcolsep}{4pt}
\begin{tabular}{llcc}
\toprule
\textbf{Dataset} & \textbf{Method} & \textbf{Spearman} & \textbf{Pearson} \\
\midrule
\multirow{6}{*}{En-De}
& CometKiwi    & 40.47 & 40.97 \\
& GEMBA        & 40.06 & 35.12 \\
& InstructScore & 35.03 & 32.85 \\
& MQMQE        & 37.88 & 25.07 \\
& DCSQE        & 43.17 & 41.64 \\
& Ours         & \textbf{47.25} & \textbf{45.84} \\
\midrule
\multirow{6}{*}{Zh-En}
& CometKiwi    & 40.35 & 35.53 \\
& GEMBA        & 33.80 & 32.56 \\
& InstructScore & 36.40 & 28.00 \\
& MQMQE        & 39.26 & 22.94 \\
& DCSQE        & \textbf{46.41} & 37.55 \\
& Ours         & 38.28 & \textbf{41.17} \\
\bottomrule
\end{tabular}
\caption{Sentence-level correlation with human MQM scores. Baseline results
are taken from \citet{geng2025alleviating}. All values are multiplied by 100.}
\label{tab:sentence-correlation}
\end{table}

Without sentence-level training, our method achieves the best Pearson
correlation on both language pairs and the best Spearman correlation on En-De,
while outperforming most baselines. These results provide complementary
sentence-level evidence, while the main evaluation remains focused on
fine-grained error localization.

\section{Inference Cost}
\label{sec:appendix-inference-cost}

Our method introduces additional routing and conditional evaluation, so its
effectiveness should be considered together with the corresponding inference
overhead. For the matched comparison, we run all four settings on the same
fixed batch of 100 translation samples under identical hardware, model,
batching, and decoding configurations. Table~\ref{tab:inference-cost}
reports the average number of model calls, input tokens, output tokens,
reasoning tokens, and wall-clock latency per item on this shared sample set.

\begin{table*}[t]
\centering
\small
\renewcommand{\arraystretch}{1.12}
\setlength{\tabcolsep}{3.5pt}
\begin{tabular}{lcccccc}
\toprule
\textbf{Setting} &
\textbf{Profiled items} &
\textbf{Calls/item} &
\textbf{Input tok./item} &
\textbf{Output tok./item} &
\textbf{Reasoning tok./item} &
\textbf{Latency (s/item)} \\
\midrule
Direct baseline & 100 & 1.000 & 207.85  & 8776.81 & 863.20  & 6.193 \\
Comprehensive   & 100 & 1.000 & 3400.00 & 6355.47 & 6158.38 & 31.270 \\
Expansion       & 100 & 2.000 & 2503.43 & 4634.94 & 4224.58 & 19.158 \\
Balanced        & 100 & 1.764 & 4523.56 & 5735.75 & 5564.53 & 25.763 \\
\bottomrule
\end{tabular}
\caption{Matched inference-cost comparison on the same fixed set of 100
samples. Token counts and latency are averages per item.}
\label{tab:inference-cost}
\end{table*}

Expansion and Balanced require approximately $3.09\times$ and $4.16\times$ the
latency of the direct baseline, respectively, while Comprehensive requires
approximately $5.05\times$. Balanced dynamic allocation remains less expensive
than applying Comprehensive evaluation to every instance, reducing average
latency from 31.270 to 25.763 seconds per item, an absolute reduction of 5.507
seconds or 17.61\%. All comparisons are computed on the same 100 samples shown
in Table~\ref{tab:inference-cost}.

The proposed framework is therefore not computationally equivalent to a
single-pass baseline. At the same time, the ablation results show that simply
exposing the largest rubric space does not consistently yield the strongest
span-level performance: Comprehensive evaluation increases recall but can
introduce substantially more false positives, whereas case-specific allocation
provides a better balance between precision and recall. Performance gains
should thus be considered together with the additional inference cost, but
cannot be attributed solely to additional computation.

\section{Additional Dynamic Rubric Complexity Analysis}
\label{sec:appendix-complexity}

We provide additional analysis of the relationship between translation
complexity and required rubric granularity. Following the main analysis, we
divide the evaluation search space into three rubric tiers:
\textit{Compact}, \textit{Expanded}, and \textit{Comprehensive}.
The additional results cover \textsc{Zh-En} with Ours-4B, \textsc{En-De} with
Ours-4B, and \textsc{En-De} with Ours-8B.

\begin{figure}[t]
    \centering
    \includegraphics[width=\linewidth]{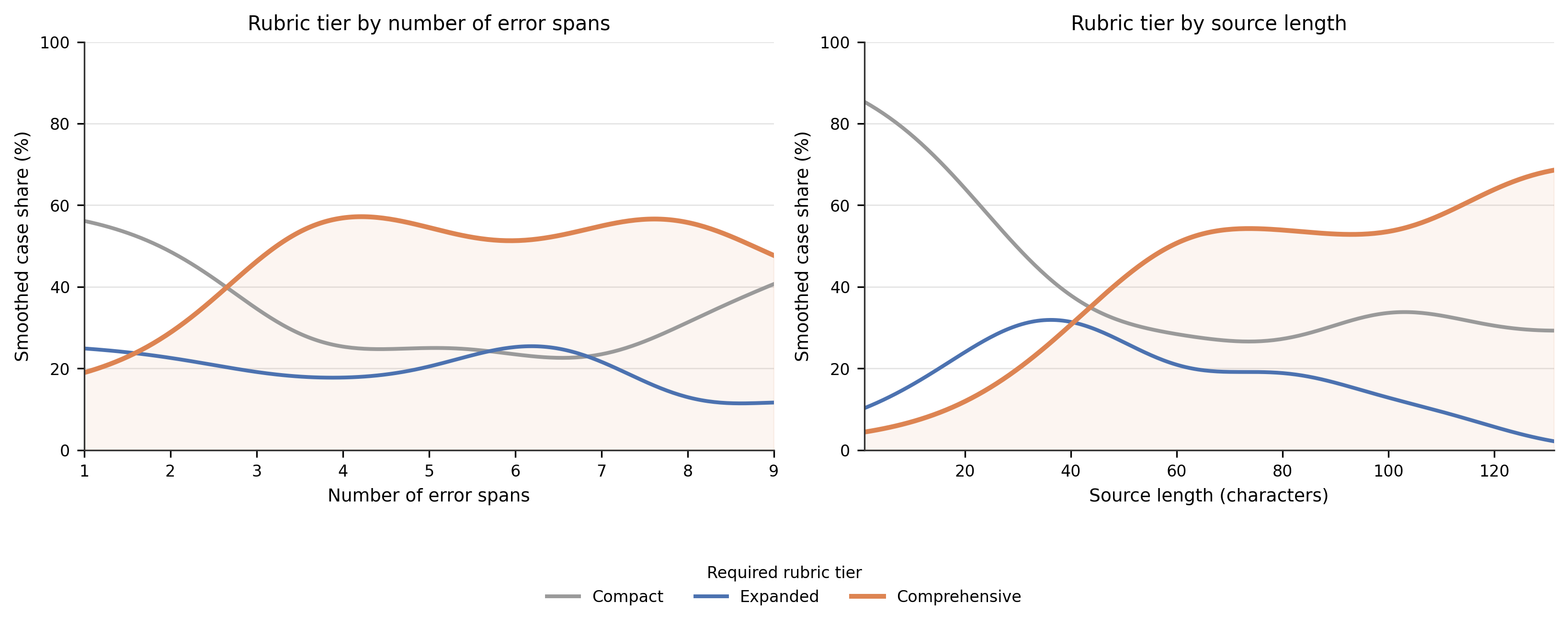}
    \caption{
    Smoothed distribution of required rubric tiers under different translation
    complexities on \textsc{Zh-En} 2023 with Ours-4B.
    Left: trend with respect to the number of error spans.
    Right: trend with respect to source sentence length.
    }
    \label{fig:app-zhen23-ours4b-rubric-trend}
\end{figure}

\begin{figure}[t]
    \centering
    \includegraphics[width=\linewidth]{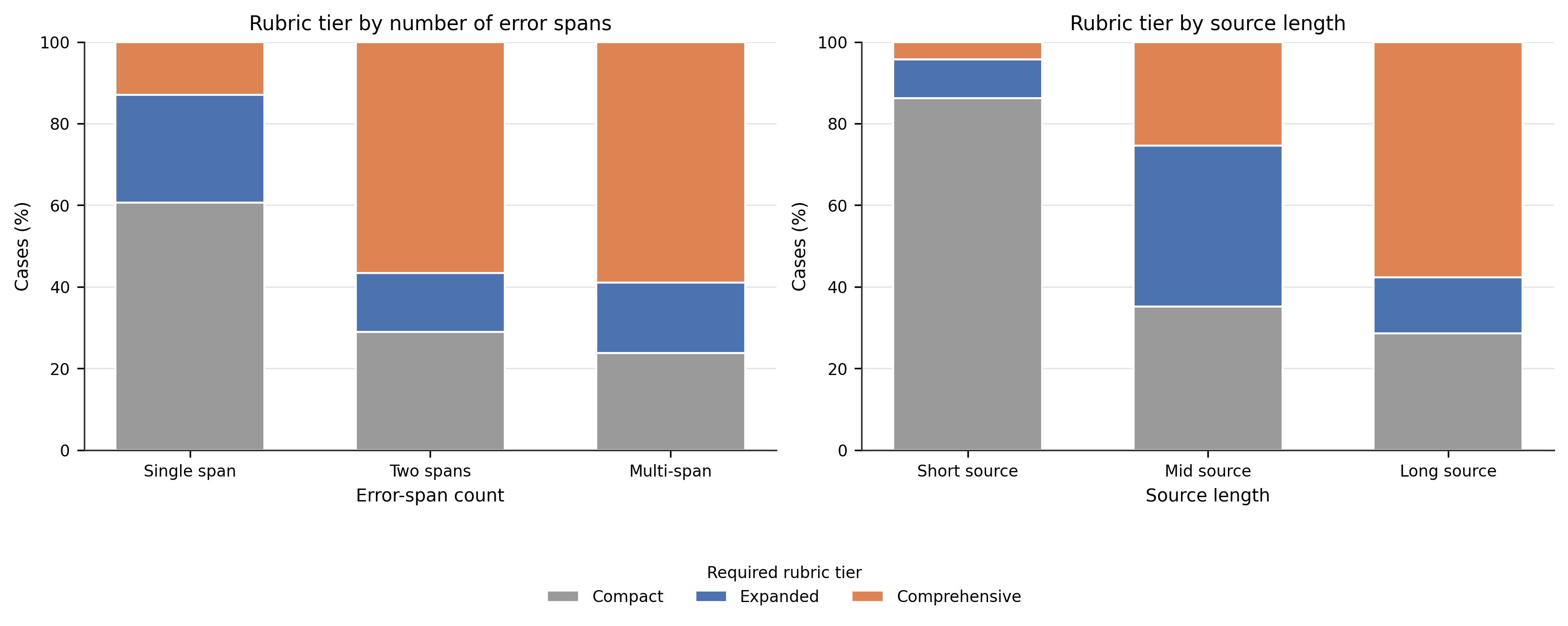}
    \caption{
    Bucketed statistics of required rubric tiers on \textsc{Zh-En} 2023 with
    Ours-4B.
    Left: grouped by error-span count.
    Right: grouped by source sentence length.
    }
    \label{fig:app-zhen23-ours4b-rubric-bucket}
\end{figure}

\begin{figure}[t]
    \centering
    \includegraphics[width=\linewidth]{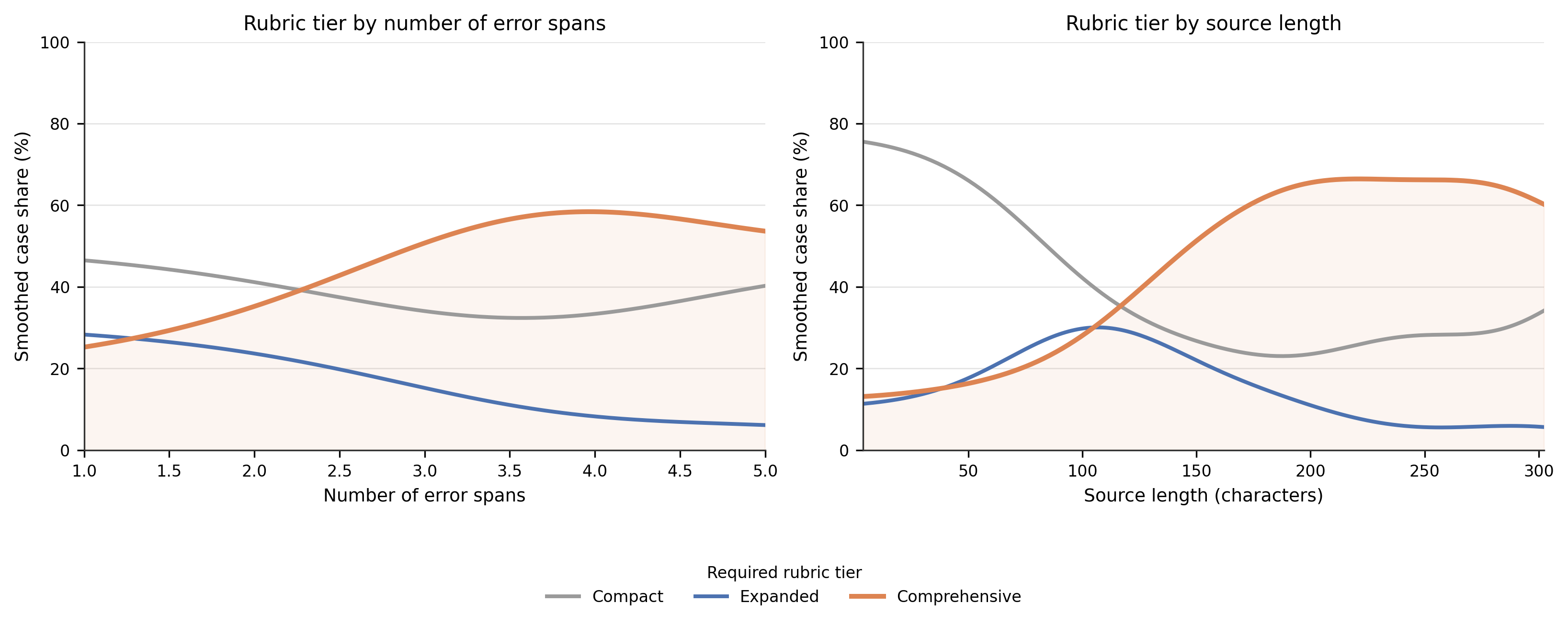}
    \caption{
    Smoothed distribution of required rubric tiers under different translation
    complexities on \textsc{En-De} 2023 with Ours-4B.
    Left: trend with respect to the number of error spans.
    Right: trend with respect to source sentence length.
    }
    \label{fig:app-ende23-ours4b-rubric-trend}
\end{figure}

\begin{figure}[t]
    \centering
    \includegraphics[width=\linewidth]{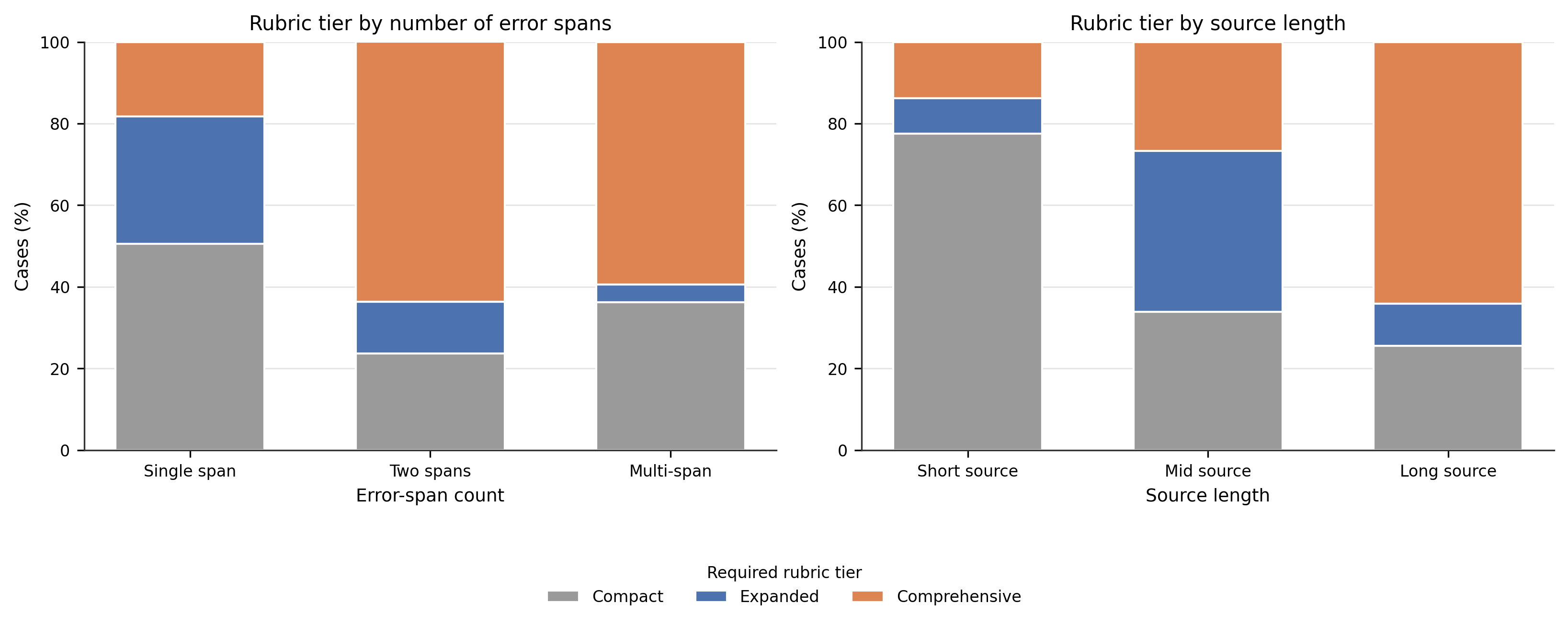}
    \caption{
    Bucketed statistics of required rubric tiers on \textsc{En-De} 2023 with
    Ours-4B.
    Left: grouped by error-span count.
    Right: grouped by source sentence length.
    }
    \label{fig:app-ende23-ours4b-rubric-bucket}
\end{figure}

\begin{figure}[t]
    \centering
    \includegraphics[width=\linewidth]{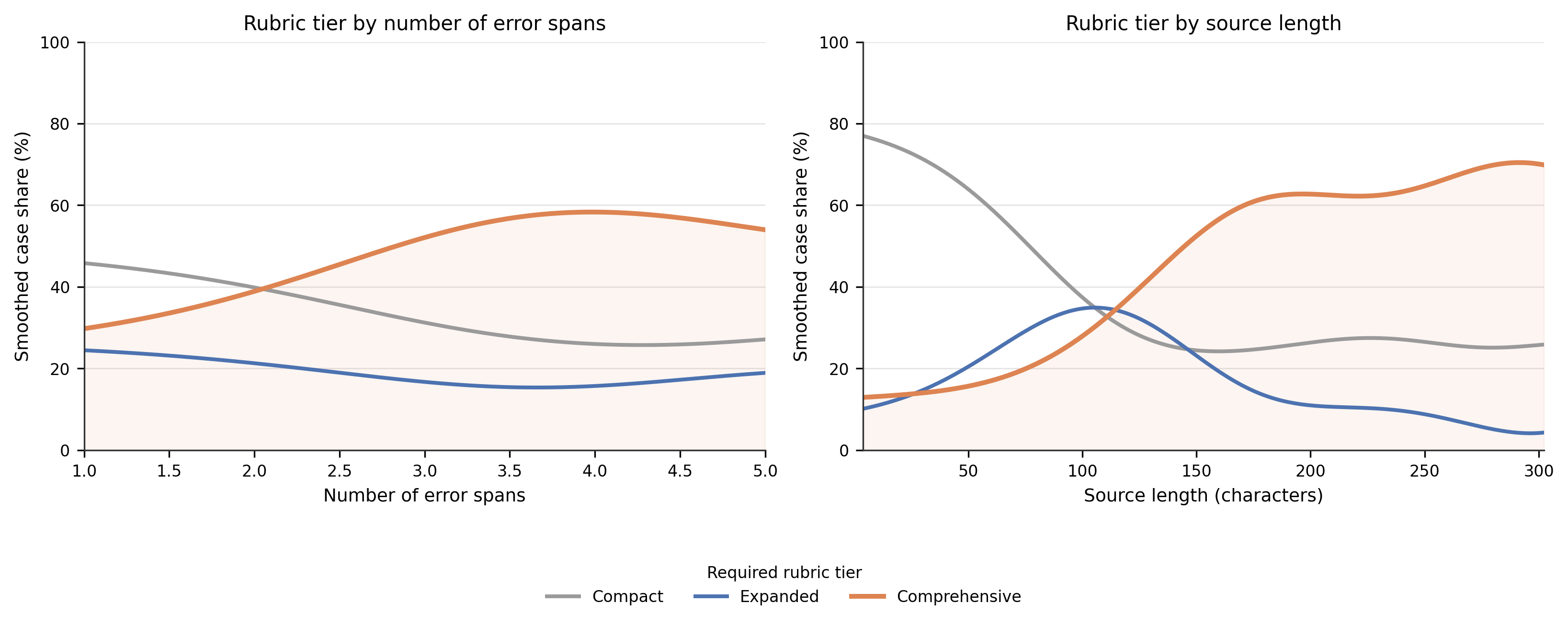}
    \caption{
    Smoothed distribution of required rubric tiers under different translation
    complexities on \textsc{En-De} 2023 with Ours-8B.
    Left: trend with respect to the number of error spans.
    Right: trend with respect to source sentence length.
    }
    \label{fig:app-ende23-ours8b-rubric-trend}
\end{figure}

\begin{figure}[t]
    \centering
    \includegraphics[width=\linewidth]{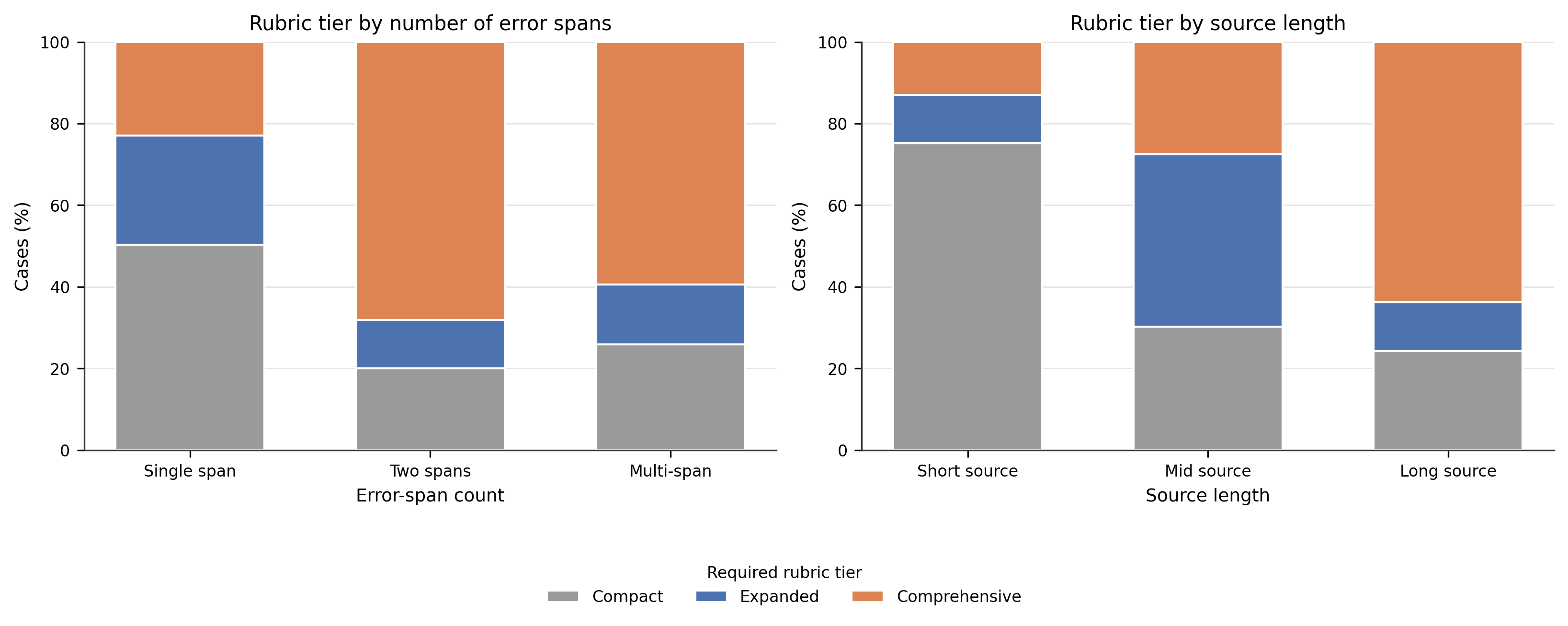}
    \caption{
    Bucketed statistics of required rubric tiers on \textsc{En-De} 2023 with
    Ours-8B.
    Left: grouped by error-span count.
    Right: grouped by source sentence length.
    }
    \label{fig:app-ende23-ours8b-rubric-bucket}
\end{figure}

Overall, the additional results show trends consistent with the main analysis.
Cases with fewer error spans or shorter source sentences are more often handled
by Compact rubric settings, while multi-span or longer inputs require larger
rubric spaces more frequently. The Expanded tier usually appears in
medium-complexity regions, serving as an intermediate level between compact and
comprehensive exploration.

These results further support the need for case-specific rubric allocation.
A fixed rubric configuration cannot equally match the evaluation needs of simple
and complex translation instances, whereas dynamic routing allows the evaluator
to adjust the exposed MQM search space according to sample complexity.
\section{Rubric Search Space Effects}
\label{sec:analysis-search-space}

Our experiments suggest that MQM rubrics function not only as explicit evaluation criteria, but also as a form of evaluation search space that influences the exploration behavior of the evaluator.

Given a source sentence $x$, translation $y$, and rubric space $\mathcal{R}$, the span prediction process can be formulated as:
\begin{equation}
    \hat{\mathcal{S}} = f(x, y, \mathcal{R}),
\end{equation}
where $\mathcal{R}$ denotes the currently activated MQM subtype space.

During free-form span prediction, the model often outputs both error spans and their associated MQM subtypes simultaneously. A natural assumption is therefore that the predicted subtypes already capture the model's internal error hypotheses, allowing the evaluation space to be compressed into:
\begin{equation}
    \mathcal{R}_{\text{compact}} \subset \mathcal{R}_{\text{full}}.
\end{equation}

Under this assumption, one may expect that retaining only the predicted subtypes should preserve most valid error spans while removing irrelevant rubric branches. In practice, however, we observe the opposite behavior. Empirical results show that aggressively restricting the rubric space substantially reduces recall. On WMT23 Zh-En using Qwen3-8B, the word-level recall drops from approximately $0.70$ to $0.46$ when only predicted subtypes are retained.

One possible explanation is that the exposed MQM criteria affect both the final labels and the evaluator's intermediate generation trajectory:
\begin{equation}
    \mathcal{R}
    \rightarrow
    \text{reasoning paths}
    \rightarrow
    \hat{\mathcal{S}}.
\end{equation}

A larger rubric space appears to encourage the model to inspect translations from multiple semantic perspectives simultaneously. Different MQM subtypes may trigger partially overlapping but distinct reasoning trajectories, including adequacy checking, fluency verification, terminology consistency inspection, stylistic comparison, or structural alignment analysis. Even when a subtype is not ultimately selected in the final output, its presence in the rubric space may still activate useful intermediate reasoning behaviors that help expose additional candidate spans.

We further observe that many valid spans discovered under larger rubric configurations cannot be reproduced after the rubric space is compressed, even when the final predicted subtype remains unchanged. This suggests that the missing spans are not caused purely by incorrect subtype prediction. Instead, restricting the rubric space alters the evaluator's decoding behavior itself and suppresses alternative exploration paths during generation.

The effect becomes particularly visible for ambiguous or weakly localized translation errors. Some spans may simultaneously admit multiple interpretations under different MQM perspectives. For example, a mistranslation may also exhibit stylistic awkwardness or terminology inconsistency. When only a narrow subset of subtypes is exposed, the evaluator tends to converge prematurely toward a limited interpretation space, reducing the probability of exploring alternative candidate spans.

This observation also helps explain why fully static large rubrics often achieve strong recall despite introducing substantial noise. Broad rubric spaces implicitly enlarge the evaluator's exploration frontier, allowing the model to retain more potentially useful hypotheses during generation. The improvement therefore does not solely originate from accurate subtype matching, but also from exposing diverse error perspectives to the evaluator.

At the same time, excessively large rubric spaces may introduce irrelevant reasoning branches and unstable predictions. The role of dynamic routing is therefore not simply to minimize the rubric space, but to allocate a sufficiently expressive evaluation space for each translation instance. Simple cases may only require compact rubric subsets, while difficult or ambiguous cases benefit from broader MQM exploration spaces.

Overall, these results suggest that MQM rubrics should be viewed as dynamic reasoning constraints rather than static label inventories. Effective span-level QE depends not only on which MQM categories are selected, but also on how the selected rubric space shapes the evaluator's exploration behavior during inference.

\end{document}